\documentclass[sigconf]{acmart}

\AtBeginDocument{%
  }

\copyrightyear{2024}
\acmYear{2024}
\setcopyright{acmlicensed}\acmConference[MM '24]{Proceedings of the 32nd ACM International Conference on Multimedia}{October 28-November 1, 2024}{Melbourne, VIC, Australia}
\acmBooktitle{Proceedings of the 32nd ACM International Conference on Multimedia (MM '24), October 28-November 1, 2024, Melbourne, VIC, Australia}
\acmDOI{10.1145/3664647.3680841}
\acmISBN{979-8-4007-0686-8/24/10}

\usepackage{balance}
\usepackage{xspace}
\usepackage{multirow}
\usepackage{graphicx}
\usepackage{subcaption}
\makeatletter
\DeclareRobustCommand\onedot{\futurelet\@let@token\@onedot}
\def\@onedot{\ifx\@let@token.\else.\null\fi\xspace}

\makeatother

\begin{document}

\title{3D-GRES: Generalized 3D Referring Expression Segmentation}

\author{Changli Wu}
\authornote{Equal contribution.}
\email{wuchangli@stu.xmu.edu.cn}
\orcid{0009-0005-3593-5142}
\affiliation{%
  \institution{Key Laboratory of Multimedia Trusted Perception and Efficient Computing, \\
  Ministry of Education of China, \\
  Xiamen University}
  \city{Xiamen}
  \state{Fujian}
  \country{China}
}

\author{Yihang Liu}
\authornotemark[1]
\email{liuyihang@stu.xmu.edu.cn}
\affiliation{%
  \institution{Key Laboratory of Multimedia Trusted Perception and Efficient Computing, \\
  Ministry of Education of China, \\
  Xiamen University}
  \city{Xiamen}
  \state{Fujian}
  \country{China}
}

\author{Jiayi Ji}
\email{jjyxmu@gmail.com}
\affiliation{%
  \institution{Key Laboratory of Multimedia Trusted Perception and Efficient Computing, \\
  Ministry of Education of China, \\
  Xiamen University}
  \city{Xiamen}
  \state{Fujian}
  \country{China}
}

\author{Yiwei Ma}
\email{yiweima@stu.xmu.edu.cn}
\affiliation{%
  \institution{Key Laboratory of Multimedia Trusted Perception and Efficient Computing, \\
  Ministry of Education of China, \\
  Xiamen University}
  \city{Xiamen}
  \state{Fujian}
  \country{China}
}

\author{Haowei Wang}
\email{asucawang@tencent.com}
\affiliation{%
  \institution{Youtu Lab, Tencent,}
  \city{Shanghai}
  \country{China}
}

\author{Gen Luo}
\email{luogen@stu.xmu.edu.cn}
\affiliation{%
  \institution{Key Laboratory of Multimedia Trusted Perception and Efficient Computing, \\
  Ministry of Education of China, \\
  Xiamen University}
  \city{Xiamen}
  \state{Fujian}
  \country{China}
}

\author{Henghui Ding}
\email{henghui.ding@gmail.com}
\affiliation{%
  \institution{Institute of Big Data, \\
  Fudan University,}
  \city{Shanghai}
  \country{China}
}

\author{Xiaoshuai Sun}
\email{xssun@xmu.edu.cn}
\affiliation{%
  \institution{Key Laboratory of Multimedia Trusted Perception and Efficient Computing, \\
  Ministry of Education of China, \\
  Xiamen University}
  \city{Xiamen}
  \state{Fujian}
  \country{China}
}

\author{Rongrong Ji}
\authornote{Corresponding author.}
\email{rrji@xmu.edu.cn}
\affiliation{%
  \institution{Key Laboratory of Multimedia Trusted Perception and Efficient Computing, \\
  Ministry of Education of China, \\
  Xiamen University}
  \city{Xiamen}
  \state{Fujian}
  \country{China}
}

\renewcommand{\shortauthors}{Changli Wu et al.}

\begin{abstract}
3D Referring Expression Segmentation (3D-RES) is dedicated to segmenting a specific instance within a 3D space based on a natural language description. However, current approaches are limited to segmenting a single target, restricting the versatility of the task. To overcome this limitation, we introduce Generalized 3D Referring Expression Segmentation (3D-GRES), which extends the capability to segment any number of instances based on natural language instructions. In addressing this broader task, we propose the Multi-Query Decoupled Interaction Network (MDIN), designed to break down multi-object segmentation tasks into simpler, individual segmentations. MDIN comprises two fundamental components: Text-driven Sparse Queries (TSQ) and Multi-object Decoupling Optimization (MDO). TSQ generates sparse point cloud features distributed over key targets as the initialization for queries. Meanwhile, MDO is tasked with assigning each target in multi-object scenarios to different queries while maintaining their semantic consistency. To adapt to this new task, we build a new dataset, namely Multi3DRes. Our comprehensive evaluations on this dataset demonstrate substantial enhancements over existing models, thus charting a new path for intricate multi-object 3D scene comprehension. The benchmark and code are available at \url{https://github.com/sosppxo/MDIN}.
\end{abstract}




\maketitle

\section{Introduction}

3D Referring Expression Segmentation (3D-RES) is a burgeoning direction in the multimodal domain, attracting widespread interest from researchers~\cite{huang2021text,wu20233d}. This task aims to segment target instances based on given natural language expressions, distinguishing itself from 3D Referring Expression Comprehension (3D-REC)~\cite{chen2020scanrefer,achlioptas2020referit3d,chen2022language,he2021transrefer3d,feng2021free,zhao20213dvg,luo20223d,ZHANG201640}, which merely locates objects with bounding boxes. 3D-RES is critical for applications in autonomous robotics, human-machine interaction, and self-driving systems, demanding not only object identification but also the generation of precise 3D masks. 

\begin{figure}
    \centering
    \includegraphics{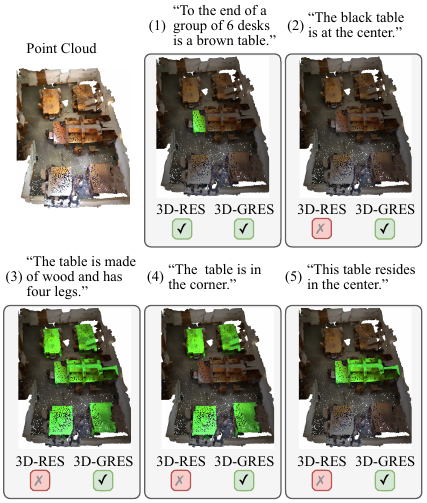}
    \caption{Traditional 3D-RES is limited to single-target cases (1). In contrast, 3D-GRES can handle scenarios with any number of targets, including no target (2), single target, and multiple targets (3-5).}
    \label{fig:1}
\end{figure}

However, traditional 3D Referring Expression Segmentation (3D-RES) settings~\cite{huang2021text,wu20233d,lin2023unified,qian2024x} are constrained to addressing single target cases, as depicted in Fig.~\ref{fig:1}-(1), a limitation significantly narrowing their practical application. In real-world scenarios, instructions often lead to situations where either no target is found or multiple targets need to be identified simultaneously, as shown in Fig.~\ref{fig:1} (2)-(5). This reality presents a challenge that existing 3D-RES models are ill-equipped to handle. To bridge this gap, we introduce a new setting, Generalized 3D Referring Expression Segmentation (3D-GRES), designed to interpret instructions specifying an arbitrary number of targets. By enhancing the Multi3DRefer~\cite{zhang2023multi3drefer} dataset through the substitution of bounding boxes with masks, we develop the Multi3DRes dataset, crafted specifically for training and validation of 3D-GRES models.

The central challenge of 3D-GRES lies in accurately identifying multiple targets from a group of similar objects. For instance, distinguishing a specific table placed in a corner from numerous similar tables, as illustrated in Fig.~\ref{fig:1}-(4). Directly applying existing 3D-RES frameworks, such as 3D-STMN~\cite{wu20233d}, to 3D-GRES tasks has proven to be ineffective. These frameworks typically employ a single query to activate the point cloud scene, culminating in a final mask. However, a single query struggles to precisely identify multiple targets among similar instances due to their identical appearances and similar semantics. 
Previous work~\cite{liu2023gres,hu2023beyond,wu2024towards,xia2023gsva,zhang2024psalm} of 2D-GRES has progressively explored this area, providing a wealth of insights. 
Inspired by this, we recognize that the key to addressing such challenges lies in decoupling the task, allowing several queries to simultaneously handle the localization of a multi-object language instruction, with each query responsible for an individual instance within the multi-object scenario. However, their decoupling methods, such as Minimap~\cite{liu2023gres}, are designed for continuous 2D images and do not extend effectively to the unordered and sparse nature of 3D point clouds.

In this paper, we introduce the Multi-Query Decoupled Interaction Network (MDIN), a novel framework specifically engineered for 3D-GRES, which is responsible for facilitating the interaction between queries and both superpoints and text. To adeptly handle an arbitrary number of targets, we incorporate a mechanism that allows multiple queries to decouple and collaboratively generate multi-object results, with each query responsible for a single target within the multi-object instance, and a classification head that determines the query-wise presence of targets. To ensure that queries evenly cover key targets in the point cloud scene, we introduce a novel Text-driven Sparse Queries (TSQ) module to generate sparse, text-related queries. Furthermore, to simultaneously achieve distinctiveness among queries and maintain overall semantic coherence, we have developed a Multi-object Decoupling Optimization (MDO) strategy. This strategy decouples the multi-object mask into individual single-object supervisions, preserving the discriminative ability of each query. By anchoring the features of the queries and the superpoint features of the ground truth in the point cloud scene to the textual semantics, it ensures semantic consistency across multiple targets.

Our contributions are threefold:
\begin{itemize}
\item We introduce a new challenging 3D-GRES task and benchmark, exploring more general and flexible interaction in 3D scenes through the development of the MDIN.
\item We design the TSQ to ensure balanced coverage of key targets within the point cloud scene and the MDO strategy to maintain the distinctiveness of queries while preserving the semantic consistency of instructions.
\item Extensive quantitative and qualitative experiments demonstrate the effectiveness of our proposed methods in addressing 3D-GRES.
\end{itemize}

\section{Related Work}

\subsection{2D Referring Expression Comprehension and Segmentation}

Research in the multimodal field~\cite{wu24next, fei2024enhancing, fei2023scene, fei2024dysen, ma2022xclip, ma2023xmesh, ma2023towards} is thriving due to its significant application value, with REC and RES tasks receiving considerable attention. The 2D-REC task aims to predict a bounding box for the target based on a referring expression~\cite{nagaraja2016modeling, hu2017modeling, yu2017joint, yu2018mattnet, deng2018visual, liu2019improving, hu2016natural, zhuang2018parallel, su2023referring, liu2022instance, liu2024remoteclip}, while 2D-RES focuses on predicting a segmentation mask that accurately represents the referred object~\cite{hu2016segmentation, luo2020multi, yang2022lavt, ye2019cross, li2018referring, liu2023caris, luo2020cascade, yang2021bottom, liu2023multi}. Common datasets for these tasks include ReferIt~\cite{2014referitgame}, RefCOCO~\cite{yu2016modeling}, and RefCOCOg~\cite{mao2016generation}, where each expression refers to a single instance.

REC and RES methods are generally categorized into one-stage~\cite{VLT, sadhu2019zero, hu2020bi, yang2020improving, wang2023unveiling, ding2021vision, jiao2021two, jing2021locate} and two-stage~\cite{wang2019neighbourhood, hong2019learning, liu2019learning, yu2018mattnet} approaches. One-stage methods fuse visual and linguistic features to directly predict segmentation masks, while two-stage methods generate instance proposals using object detection or segmentation and then use language features to select the target instances. Despite their success in the 2D domain, these methods face challenges in the 3D domain due to the irregularity and sparsity of 3D point clouds.

\subsection{3D Referring Expression Comprehension and Segmentation}

With the widespread adoption of deep learning techniques in 3D point clouds, the 3D-REC task has attracted significant attention. \citet{chen2020scanrefer} released the ScanRefer dataset for the 3D visual grounding task, which involves locating referred objects in 3D scenes. Referit3d~\cite{achlioptas2020referit3d} proposes two datasets, Nr3d and Sr3d, where referred objects belong to fine-grained object categories, and scenes contain multiple instances of such categories.
While most existing methods adopt a two-stage~\cite{chen2022language,he2021transrefer3d,feng2021free,zhao20213dvg,yuan2021instancerefer} paradigm, some researchers have also explored one-stage~\cite{luo20223d,wang20233drp} approaches.

In contrast, research on 3D-RES~\cite{huang2021text, wu20233d, he2024refmask3d, he2024segpoint} is still in its nascent stage. TGNN~\cite{huang2021text}, as a pioneer in the 3D-RES task, proposed a two-stage method based on Graph Neural Networks. 3D-STMN~\cite{wu20233d} introduced a one-stage approach, significantly improving inference speed. 
These methods mainly focus on single-target descriptions, whereas our work is primarily dedicated to segmenting a flexible number of target objects based on language descriptions.

\subsection{Generalized Referring Expression Comprehension and Segmentation }
The original 2D-REC and RES tasks do not specify a limit on the number of target instances. However, previous datasets~\cite{2014referitgame,yu2016modeling,mao2016generation} typically assume each expression refers to a single target, which creates issues when no or multiple targets are present. To address this, \citet{liu2023gres} introduced the Generalized Referring Expression Segmentation (GRES) benchmark to handle expressions referring to zero, single, or multiple targets. \citet{GREC} further developed this approach with Generalized Referring Expression Comprehension (GREC). Subsequent research~\cite{wu2024towards,hu2023beyond,dang2023instructdet,xie2024described,huang2023dense} has expanded 2D-REC and RES tasks to include multiple target descriptions, with methods~\cite{xia2023gsva,zhang2024psalm,zhao2024open} showing strong performance on GRES.

In the 3D domain, research is limited. \citet{zhang2023multi3drefer} extended the ScanRefer task to Multi3DRefer, which locates a variable number of targets in 3D scenes using bounding boxes. However, bounding boxes can be ambiguous in dense scenes. We propose the 3D Generalized Referring Expression Segmentation (3D-GRES) task to improve localization by generating segmentation masks based on language descriptions.

\begin{figure*}
    \centering
    \includegraphics[width=0.9\textwidth]{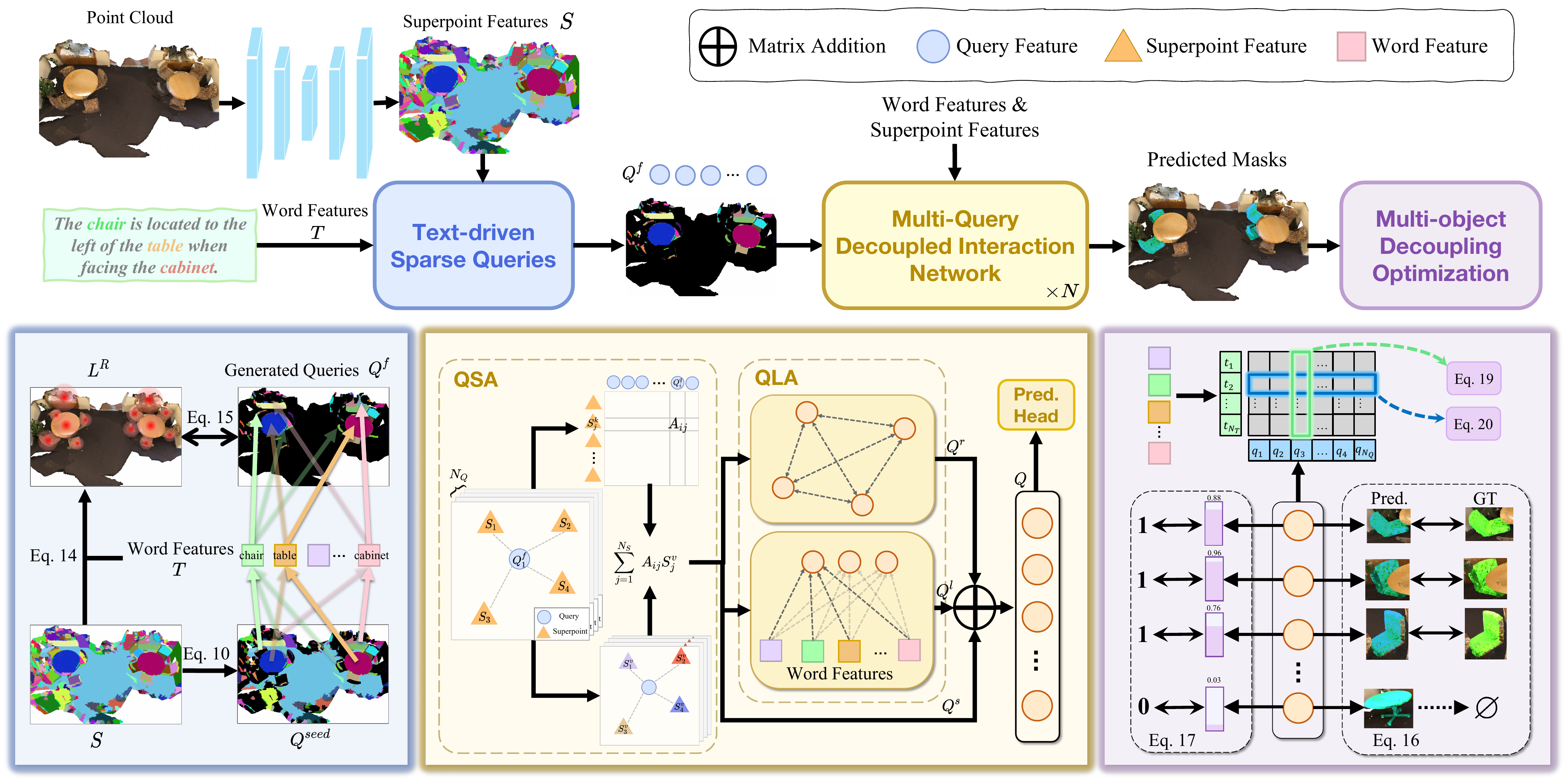}
    \caption{The overall framework of MDIN, comprising its core modules TSQ and MDO. The input point cloud and text undergo feature extraction before being fed into the TSQ module to extract sparse decoupled queries. Subsequently, the MDIN module performs multimodal fusion and prediction. Finally, the MDO module carries out decoupled optimization.}
    \label{fig:2}
\end{figure*}

\section{3D-GRES}
\subsection{Classical 3D-RES}
The classic 3D-RES task~\cite{huang2021text,wu20233d} is focused on generating a 3D mask for a single target object within a point cloud scene, guided by a referring expression. This traditional task exhibits significant limitations.
Firstly, it fails to accommodate scenarios where no object within the point cloud scene matches the given expression. An illustrative example of this is depicted in Fig.~\ref{fig:1} (2), where no object corresponds to the expression ``the black table is at the center''.
Secondly, it does not account for instances where multiple objects fit the described criteria. For example, in Fig.~\ref{fig:1} (4), the expression ``the table is in the corner'' applies to four distinct objects.
This significant gap between model capabilities and real-world applicability restricts the practical deployment of 3D-RES technologies in scenarios reflective of everyday complexities.

\subsection{3D-GRES Settings}
\noindent \textbf{Settings and Benchmark}. To overcome existing limitations, we introduce the Generalized 3D Referring Expression Segmentation (3D-GRES) task, designed to identify an arbitrary number of objects from textual descriptions. 3D-GRES involves processing a 3D point cloud scene $P$, a referring textual expression $E$, and generating a corresponding 3D mask $M$, which can signify zero, one, or multiple objects. For expressions indicating no targets, $M$ will be an all-zero mask, showing no selected points in $P$. It enables locating multiple objects through multi-target expressions and verifying the existence of specific objects in a scene with ``nothing'' expressions, thus offering enhanced flexibility and robustness in object retrieval and interaction within 3D environments.

\noindent \textbf{Multi3DRes Dataset}.
To adapt to the 3D-GRES task, we utilize the Multi3DRefer dataset~\cite{zhang2023multi3drefer} to create the Multi3DRes dataset. The original Multi3DRefer dataset includes 61,926 language expressions referring to 11,609 objects across 800 ScanNet~\cite{dai2017scannet} scenes, with 6,688 expressions matching zero targets and 13,178 matching multiple targets. However, Multi3DRefer was designed for referring expression comprehension, yielding bounding boxes. To enhance precision, we construct the Multi3DRes dataset using instance masks from ScanNet~\cite{dai2017scannet}. This dataset extends the original by incorporating samples with no targets and multiple targets, thereby increasing task complexity and enabling more robust handling of real-world free-form user inputs.

\noindent \textbf{Metrics}.
As for evaluation metrics, we employ the conventional mIoU (Mean Intersection over Union) and introduce Acc@$k$IoU, which measures the proportion of predicted masks with an IoU greater than $k$ compared to the ground truth masks, with $k$ belonging to the set \{0.25, 0.5\}. In cases where no target objects are present, we assume an IoU of 1 for correct predictions and an IoU of 0 for incorrect predictions. Specifically, for our proposed query decoupling method, MDIN, we adjust the IoU values of individual samples based on the predicted query-wise confidence $A^{tgt}$ indicating the presence of a target. If all queries are correctly predicted as zero-target (\textit{i.e.,} $A^{tgt}>0.5$ is a zero vector), their IoU is set to 1; otherwise, it is set to 0.

Furthermore, we categorize the samples into five classes based on the Multi3DRefer~\cite{zhang2023multi3drefer} manner: a) zero target without distractors of the same semantic class; b) zero target with distractors; c) single target without distractors; d) single target with distractors; and e) multiple targets. 
Notably, classes c) and d) align with the "unique" and "multiple" cases, respectively, as defined in ScanRefer~\cite{chen2020scanrefer}. Classes a) and c) represent easier scenarios where either zero or a single target object of its semantic class is present in the scene, while classes b) and d) represent more challenging situations involving one or multiple target objects of the same semantic class in the scene.

\section{Method}

\subsection{Feature Extraction}
\subsubsection{Linguistic Feature and Text Decoupling}\label{sec:lan_feat}
In our approach, we first process the input referring expressions by encoding them into text tokens $\mathcal{T}\in\mathbb{R}^{N_T\times D_T}$, utilizing the pre-trained RoBERTa~\cite{liu2019roberta}. Here, $N_T$ represents the total number of tokens, and $D_T$ indicates the dimensionality of the original text features. To facilitate multimodal alignment, these encoded features are then mapped to a multimodal space of dimension $D$. This mapping is achieved through a linear transformation, denoted as $T = \mathcal{T}W_T$, where $T\in\mathbb{R}^{N_T\times D}$ represents the projected linguistic features, and $W_T\in\mathbb{R}^{D_T\times D}$ is a learnable parameter matrix.

Following \citet{wu2023eda}, we use off-the-shelf natural language processing tools~\cite{schuster2015generating,wu2019unified} to parse descriptions into five semantic components: the \texttt{Main object} (the primary subject), the \texttt{Auxiliary object} (which helps locate the main object), \texttt{Attributes} (describing appearance and shape), \texttt{Pronoun} (substituting for the main object), and \texttt{Relationship} (the spatial relation between main and auxiliary objects). This breakdown facilitates a deeper understanding of entity relationships.

Next, we create position labels $L_{main}$, $L_{attri}$, $L_{auxi}$, $L_{pron}$, and $L_{rel} \in \mathbb{R}^{1 \times N_T}$ for words associated with each component, where $N_T$ represents the maximum text length, consistent with~\cite{wu2023eda}. In these labels, the position of words about each component is marked with a 1, while all other positions are set to 0. To obtain the text feature of a decoupled component $t\in\mathbb{R}^{1\times D}$, we perform a dot multiplication of the position label $L\in \mathbb{R}^{1 \times N_T}$ with the feature matrix of all words $T\in \mathbb{R}^{N_T\times D}$: 
\begin{equation}
    t = L\cdot T,
\end{equation}
where $t_{main}=L_{main}\cdot T$, $t_{attri}=L_{attri}\cdot T$ and $t_{pron}=L_{pron}\cdot T$ are selected to form the positive word features $T^+$ for the target.

\subsubsection{Visual Feature and Superpoint Pooling}
For the input point cloud with positions $P^P\in\mathbb{R}^{N_P\times 3}$ and RGB $C_P\in\mathbb{R}^{N_P\times 3}$, the point-wise features $F^{point}\in \mathbb{R}^{N_P\times D_P}$ can be extracted through a Sparse 3D U-Net~\cite{graham20183d}, where $N_P$ denotes the number of points and $D_P$ represents the original visual feature dimension. And we project $F^{point}$ into $D$-dimensional multimodal space: $F^P = F^{point}W_P$, where $F^P\in\mathbb{R}^{N_P\times D}$ is projected point feature and $W_P\in\mathbb{R}^{D_P\times D}$ is a learnable parameter. 
Then, we follow~\citet{sun2023superpoint,wu20233d} to obtain $N_S$ superpoints $\{K_i\}^{N_S}_{i=1}$~\cite{landrieu2018large} from the original point cloud. Finally, we directly feed point-wise features $F^P$ into superpoint pooling layer~\cite{sun2023superpoint} based on $\{K_i\}^{N_S}_{i=1}$ to get the superpoint-level features $S \in \mathbb{R}^{N_S\times D}$.

\subsection{Multi-Query Decoupled Interaction Network}\label{sec:MDIN}
Prior work like 3D-STMN~\cite{wu20233d} used a single query to activate a mask on a point cloud scene, effectively localizing a single target in 3D-RES tasks. However, this approach struggles with multiple or unspecified targets. To address this, we introduce the Multi-Query Decoupled Interaction Network (MDIN) for 3D-GRES, inspired by~\citet{liu2023gres}. MDIN uses multiple queries to handle individual instances in multi-target scenarios, aggregating these into a final result. For scenes without defined targets, predictions are made based on the confidence scores of each query, with zero targets predicted if all queries score low.

MDIN comprises multiple identical modules in series, each consisting of Query-Superpoint Aggregation (QSA) and Query-Language Aggregation (QLA) modules, facilitating interactions between queries, superpoints, and text. Unlike existing models like DETR~\cite{carion2020detr} that use random initialization, MDIN employs a Text-driven Sparse Queries (TSQ) module to generate sparse queries \( Q^f \in \mathbb{R}^{N_Q \times D} \) from \( S \), ensuring effective scene coverage, as detailed in Sec.~\ref{sec:sparseQueries}. To support multiple queries, we implement a Multi-object Decoupling Optimization (MDO) strategy to refine performance, detailed in Sec.~\ref{sec:align}.

\subsubsection{Query-Superpoint Aggregation (QSA)}
The query $Q^f$ can be considered as an anchor in the point cloud scene~\cite{shi2020points}. By enabling queries to interact with superpoints, queries can capture the global information of a point cloud scene. Notably, the sampled superpoints serve as queries during this interaction process, allowing for stronger local aggregation. This local focus creates favorable conditions for the decoupling of queries. The architecture is depicted in Fig.~\ref{fig:2}. Initially, the similarity distribution is computed between the superpoint feature $S$ and query embeddings $Q^f$:
\begin{equation} \label{eq:general_attn}
    \begin{split}
        Q^q=Q^f W_{sq}, \quad S^k= S W_{sk},\\
        A_{ij}= \frac{{\rm Sim}{\left(Q^q_i,S^k_j\right)}}{\sum_{j=1}^{N_S}\ {\rm Sim}{\left(Q^q_i,S^k_j\right)}},
    \end{split}
\end{equation}
where $W_{sq}$ and $W_{sk}$ are learnable $D\times D$ parameters and $ {\rm Sim}{\left(\cdot,\cdot \right)} $ represents the similarity function, which in this case is defined as $ {\rm Sim}\left(q,k\right)\!=\!{\rm exp}({q\cdot k^T}/{\sqrt D}) $.
Subsequently, the queries will aggregate their respective related superpoints using these similarity distributions: 
\begin{equation}
    \begin{split}
        S^v &= SW_{sv}, \\
        Q^s_i &= \sum_{j=1}^{N_S}\ A_{ij} S^v_j,
    \end{split}
\end{equation}
where $S\in \mathbb{R}^{N_S\times D}$ denotes the features of the superpoints, $W_{sv}\in \mathbb{R}^{D\times D}$ represents learnable parameters. 
Subsequently, the updated scene-aware $Q^s$ is fed into QLA for additional language-aware interactions.

\subsubsection{Query-Language Aggregation (QLA)}
The scene-aware queries $Q^s$ come from collating superpoint features that do not contain relationships between queries and language information. We propose QLA module to model the query-query and query-language interactions. As in Fig.~\ref{fig:2}, QLA consists of a self-attention\cite{vaswani2017attention} for query features $Q^s$, and a multi-modal cross attention. The self-attention models the query-query dependency relationships. It computes the attention matrix by interacting one query with all other queries and outputs the relation-aware query feature $Q^r$:
\begin{equation}
    Q^r= \text{Softmax}(Q^sW_{qq}(Q^sW_{qk})^T) \cdot 
 Q^sW_{qv},
\end{equation}
where $W_{qq},W_{qk},W_{qv}\in \mathbb{R}^{D\times D}$ are learnable parameters.
Meanwhile, the query-language interactions follow the cross attention way, and firstly models
the relationship between each word and each query:
\begin{equation}
    A^l = \text{Softmax}(Q^sW_{lq}(TW_{lk})^T),
\end{equation}
where $A^l\in \mathbb{R}^{N_Q\times N_T}$ and $W_{lq},W_{lk}\in\mathbb{R^{D\times D}}$ are learnable parameters. Then it forms the language-aware query features using the derived word-query attention: 
\begin{equation}
    Q^l=A^lT. 
\end{equation}

Finally, the relation-aware query feature $Q^r$, language-aware query features $Q^l$, and scene-aware query features $Q^s$ are added together fused by an MLP: 
\begin{equation}
    Q = \text{MLP}(Q^s + Q^r + Q^l).
\end{equation}

\subsubsection{Prediction Head}
For the $n$-th query, we first get its corresponding mask $M_n\in\{0,1\}^{N_S}$ by multiplying its feature $Q_n$ with the mask branch of superpint features $S_M$. Meanwhile, $Q_n$ is also fed into an MLP to predict a confidence $A^{tgt}_n$ that indicates its probability of containing targets. We get the final predicted mask $\mathcal{M}$ by logically aggregating these masks:
\begin{equation}
\begin{split}
    &S_M = SW_M,\quad M_n = \text{Binarization}(S_M \cdot Q_n),\\
    &\mathcal{M} = \text{Binarization}\left(\sum_n\ \mathbb{I}(A^{tgt}_n>0.5)\cdot M_n\right),
\end{split}
\end{equation}
where $S\in \mathbb{R}^{N_S \times D}$ is the superpoint features, $W_M\in \mathbb{R}^{D\times D}$ is learnable parameters, $\text{Binarization}(\cdot)$ denotes the binarization operation, which returns 1 when the input is a positive value and 0 otherwise, and $\mathbb{I}(A^{tgt}_n>0.5)$ denotes whether the $n$-th query contains one of the target instances, which can be easily supervised by whether the query belongs to a particular GroundTruth instance due to its visual-based generation process.

\subsection{Text-driven Sparse Queries\label{sec:sparseQueries}}
Directly using text tokens as queries~\cite{wu20233d} is suboptimal due to the entanglement of semantic descriptions in the text, making it difficult to differentiate between multiple objects. On the other hand, employing text-agnostic learnable parameters as queries~\cite{sun2023superpoint,lai2023mask} requires a larger quantity to cover the space adequately and relies on unstable Hungarian matching, leading to difficulties in convergence and inefficient performance.  Therefore, we propose a Text-driven Sparse Queries module to achieve sparse linguistic-aware query generation and facilitate a natural correspondence between queries and visual instances, paving the way for decoupling.

\subsubsection{The Generation Process}
In order to achieve a sparse distribution of initialized queries within the point cloud scene while preserving geometric and semantic information to a greater extent, we adopt the technique of farthest point sampling~\cite{moenning2003fast} directly for superpoints, namely the strategy of Farthest Superpoint Sampling (FSS):
\begin{equation}
     P^S_i = \text{Pool}(K_i,P^P),\ \text{FSS}(P^P) = \text{FPS}(P^S),
\end{equation}
where $P^P\in\mathbb{R}^{N_P\times 3}$ denotes the positions of points in the original point cloud, $P^S\in\mathbb{R}^{N_S\times 3}$ denotes the positions of the superpoints, $K_i$ denotes the $i$-th superpoint's mask, $\text{Pool}(\cdot)$ is the operation of average pooling and $\text{FPS}(\cdot)$ is the operation of farthest point sampling~\cite{moenning2003fast}, which returns the indices of the sampled points.

Subsequently, we utilize the features of superpoints obtained through Farthest superpoint sampling as initial seed queries, which participate in the subsequent stage of linguistic-aware refinement:
\begin{equation}
    Q^{seed}=S[\text{FSS}(P^P)],
     P^{seed}=P^S[\text{FSS}(P^P)],
\end{equation}
where $Q^{seed}\in\mathbb{R}^{N_{seed}\times D}$, $P^{seed}\in\mathbb{R}^{N_{seed}\times 3}$ denotes the features and the positions of seed queries, $N_{seed}$ denotes  the total number of seed queries, $S$ denotes the features of superpoints and the $[\cdot]$ means the operation of indexing.

The $Q^{seed}$ obtained through farthest point sampling is linguistic-agnostic, tending towards comprehensive scene coverage~\cite{cheng2021back,liu2021group,qi2019deep}. However, the referring expressions often relate only to a subset of objects. Consequently, using $Q^{seed}$ directly as a query results in significant redundancy and interferes with the accurate determination of the target objects. Therefore, we further refined the selection of $Q^{seed}$ by ranking them based on their average correlation scores with the referring description features $T$, retaining the top $N_Q$ superpoints most relevant to the referring description:
\begin{equation}
    R_i = \text{Average}(T\cdot Q^{seed}_i),
\end{equation}
\begin{equation}
    Q^f = Q^{seed}[\text{ArgTopk}(R, N_Q)],
\end{equation}
where $R_i\in \mathbb{R}$ denotes the average correlation scores of the $i$-th seed query, $T\in \mathbb{R}^{N_T\times D}$ denotes the linguistic features, $Q^{seed}_i$ denotes the feature of the $i$-th seed query, $T\cdot Q^{seed}_i$ denotes the similarity matrix between $T$ and $Q^{seed}_i$, $\text{Average}(\cdot)$ denotes the averaging operation, $R\in \mathbb{R}^{N_{seed}}$ denotes the correlation scores of seed queries, $\text{ArgTopk}(\cdot, N_Q)$ returns the indices of the top $N_Q$ elements with the highest values and $Q^f\in \mathbb{R}^{N_Q\times D}$ denotes the finally selected superpoint features, which will serve as the queries.

\subsubsection{Query Generation Decoupling Loss} 
To supervise the relevance scores of descriptions, a common approach is to employ the mentioned categories as labels for the seed queries. However, this simplistic approach often leads to queries that overly prioritize easily identifiable objects, resulting in the omission of other mentioned objects in the sampling process.

Therefore, we introduce Query Generation Decoupling Loss, which is categorized into three scenarios: Taking the $n$-th  mentioned instance as an example, the label of the seed query closest to the center of the $n$-th mentioned instance is set to 1; for the remaining seed queries belonging to the $n$-th mentioned instance, their labels are determined using a Gaussian distribution based on their distances to the center of  the $n$-th mentioned instance; the labels of seed queries not belonging to any instance are set to 0: 
\begin{equation}
    Q_{n} =\text{ArgTop1}(\text{Dist}(Q^{seed}, I^{mentioned}_n)),
\end{equation}

\begin{equation}
    L^R_i =\left\{
    \begin{aligned}
    1 &\quad \text{if}\quad Q^{seed}_i = Q^{seed}_n, \\
    \exp(-\alpha  \cdot \frac{dist_{i,n}^2}{\sigma^2})  &\quad \text{if}\quad Q^{seed}_i \in I^{mentioned}_n, Q^{seed}_i \neq Q^{seed}_n,  \\
    0  &\quad \text{otherwise},
    \end{aligned}
    \right.
\end{equation}
where $Q_n$ denotes the seed query nearest to the center of the $n$-th mentioned instance, $I^{mentioned}_n$ denotes the $n$-th mentioned instance, $L^R_i$ denotes the relevance label of the $i$-th seed query, $Q_i$ denotes the $i$-th seed query, $dist_{i,n}$ denotes the distance from the $i$-th seed query to the center of the $n$-th mentioned instance, $\alpha$ is a control factor used to adjust the peak and shape of the Gaussian distribution, $\sigma$ is the standard deviation of the Gaussian distribution. Finally, the instance decoupling guidance loss is defined as:
    
\begin{equation}
    \mathcal{L}_{qgd} = \text{BCE}(R,L^R),
\end{equation}
where $R, L^{R}\in \mathbb{R}^{N_{seed}}$ denote the predictions and labels of the correlation scores for the seed queries.

\begin{table*}
\renewcommand{\arraystretch}{1.2}
\centering
\caption{Results of 3D-GRES task on Multi3DRes, where ``zt w/ dis'' means zero target with distractor, ``zt w/o dis'' means without distractor, ``st w/ dis'' means single target with distractor, ``st w/o dis'' means without distractor, ``mt'' means multiple target.}
\resizebox{1.0\linewidth}{!}{
\begin{tabular}{l|c|cccccc|cccccc} 
\toprule
 &
   \multirow{2}{*}{\centering \textbf{mIoU}} & \multicolumn{6}{c|}{\textbf{Acc@0.25}} & 
   \multicolumn{6}{c}{\textbf{Acc@0.5}}  \\
  \multirow{-2}{*}{\centering \textbf{Method}}  &
  &\textbf{zt w/ dis} &\textbf{zt w/o dis} &\textbf{st w/ dis} 
  &\textbf{st w/o dis}  &\textbf{mt}  &\textbf{Overall} 
  &\textbf{zt w/ dis}  &\textbf{zt w/o dis} &\textbf{st w/ dis}  &\textbf{st w/o dis} &\textbf{mt}  &\textbf{Overall}\\
\midrule
\textbf{ReLA~\cite{liu2023gres}}
& 42.8 & 36.2 & 72.7 & 48.3 & 83.4 & 73.0 & 61.8 & 36.2 & 72.7 & 20.4 & 65.5 & 42.4 & 37.4\\
\textbf{M3DRef-CLIP~\cite{zhang2023multi3drefer}}
& 37.4 & 39.2 & \textbf{81.6} & 50.8 & 77.5 & 66.8 & 55.7 & 39.2 & \textbf{81.6} & 29.4 & 67.4 & 41.0 & 37.5\\
\textbf{3D-STMN~\cite{wu20233d}} 
& 43.0 & 42.6 & 76.2 & 49.0 & 77.8 & 68.8 & 60.4 & 42.6 & 76.2 & 24.6 & 69.2 & 43.9 & 40.9\\
\textbf{Ours}
& \textbf{47.5} & \textbf{47.9} & 78.8 & \textbf{55.5} & \textbf{84.4} & \textbf{76.3} & \textbf{67.0} & \textbf{47.9} & 78.8 & \textbf{29.5} & \textbf{71.7} & \textbf{46.8} & \textbf{44.7}\\
\bottomrule
\end{tabular}
}
\label{3dgres}
\end{table*}

\subsection{Multi-object Decoupling Optimization}\label{sec:align}

\subsubsection{Decoupling Mask Loss}

To refine the segmentation capability of each query, we continue to leverage the intrinsic attributes of queries generated by TSQ, where each query corresponds to an object within the point cloud. Building upon this foundation, we assign each query the responsibility of predicting the mask for its corresponding GroundTruth object.
Specificly, for the $n$-th target instance GT mask $M^{tgt}_n\in \mathbb{R}^{N_S}$, we filter out the corresponding query $Q^{+}_n\in\mathbb{R}^{D}$ and get the mask loss:
\begin{equation}
    \mathcal{L}_{mask} = \text{BCE}(SW_M\cdot Q^{+}_n, M^{tgt}_n) + \text{DICE}(SW_M\cdot Q^{+}_n, M^{tgt}_n),
\end{equation}
where $S \in \mathbb{R}^{N_S\times D}$ denotes the superpoint features, $W_M\in \mathbb{R}^{D\times D}$ is learnable parameters, $\text{BCE}$ is the binary cross-entropy loss and $\text{DICE}$ is the Dice loss~\cite{milletari2016v}. $Q^{+}$ can be easily obtained by determining whether the query belongs to one of the target instances.

\subsubsection{Target Confidence Decoupling Loss}

For the $n$-th instance, we predicted the confidence score $A^{tgt}_n$ indicating the presence of the target object using the MDIN approach described in Section 4.2. Leveraging the generation method of our Text-driven Sparse Queries Generation, each query corresponds directly to the instance $I^n_{tgt}$, obtained through instance labels. 

Consequently, we can construct labels $L^{tgt} \in \{0,1\}^{N_Q}$ indicating whether each query contains the target object, where 1 represents the presence of the target and 0 represents its absence. Thus, we employ the binary cross-entropy (BCE) loss to enable each query to learn its discriminative ability regarding whether it belongs to the target object: \begin{equation}
    \mathcal{L}_{tgt} = \text{BCE}(A^{tgt}, L^{tgt}),
\end{equation}
where $A^{tgt}\in \mathbb{R}^{N_Q}$ is the predicted confidence of queries and $L^{tgt}\in \{0,1\}^{N_Q}$ denotes the corresponding label indicating whether the query contains one of the target instances.

\subsubsection{Query-Text Alignment Loss}
To enhance the decoupling of queries and delegate them to individual instances, we leverage the intrinsic attributes of queries generated by TSQ. Each query originates from a superpoint within the point cloud, inherently associating it with a specific object. Queries for GroundTruth target instances are responsible for segmenting their corresponding instances, while unassociated instances are assigned to the nearest query. This method uses prior visual constraints to disentangle queries and assign them to individual instances.

Inspired by~\citet{wu2023eda}, we use the discriminative semantic features of referring text to align queries for GroundTruth instances with positive textual features \( T^+ \) (i.e., word features corresponding to target objects as described in Sec.~\ref{sec:lan_feat}), while pushing other queries away. This is achieved through a decoupled contrastive learning approach comprising query-word loss and word-query loss, defined as follows:
\begin{small}
\begin{equation}
    q=QW_{q},\quad t=TW_{w},
\end{equation}
\begin{equation}
    \mathcal{L}_{q\rightarrow w}\! =\! \sum_{i=1}^{N_Q} \frac{1}{\left|\mathbf{T}_{i}^{+}\right|} \!\sum_{{t}_i \in \mathbf{T}_{i}^{+}}\!\!\!-\log \!\left(\!\frac{\exp \left( ({q}_{i}^{\top} {t}_{i} / \tau)\right)}{\sum_{j=1}^{N_T} \exp \left(({q}_{i}^{\top} {t}_{j} / \tau)\right)}\!\right),
    \label{eq:loss_qta_qurey}
\end{equation}
\end{small}
\begin{small}
\begin{equation}
    \mathcal{L}_{w\rightarrow q} = \sum_{i=1}^{N_T} \frac{1}{\left|\mathbf{Q}_{i}^{+}\right|} \sum_{{q}_i \in \mathbf{Q}_{i}^{+}}-\log \left(\frac{\exp \left ({t}_{i}^{\top} {q}_{i} / \tau\right)}{\sum_{j=1}^{N_Q} \exp \left ({t}_{i}^{\top} {q}_{j} / \tau \right)}\right),
    \label{eq:loss_qta_text}
\end{equation}
\end{small}
where $q\in \mathbb{R}^{N_Q\times C},t\in \mathbb{R}^{N_T \times C}$ are the query and word features, $W_q, W_w\in \mathbb{R}^{D\times C}$ are learnable parameters.
$N_Q$ and $N_T$ are the number of queries and words. $t_i \in T^+_i$ denotes the positive word feature of the $i$-th query and $q_i \in Q^+_i$ denotes the positive query feature corresponding to the $i$-th word. And we get the final Query-Text Alignment Loss $\mathcal{L}_{qta}$ by average $\mathcal{L}_{q\rightarrow w}$ and $\mathcal{L}_{w\rightarrow q}$:
\begin{equation}
    \mathcal{L}_{qta} = \mathcal{L}_{q\rightarrow w} + \mathcal{L}_{w\rightarrow q}.
\end{equation}

The final loss is calculated as the weighted sum of 
$\mathcal{L}_{qgd}$, $\mathcal{L}_{mask}$, $\mathcal{L}_{tgt}$ and $\mathcal{L}_{qta}$ :
\begin{equation}
    \mathcal{L} = \lambda_{qgd}\mathcal{L}_{qgd} + \lambda_{mask}\mathcal{L}_{mask} + \lambda_{tgt}\mathcal{L}_{tgt} + \lambda_{qta}\mathcal{L}_{qta},
\end{equation}
where $\lambda_{qgd}$, $\lambda_{mask}$, $\lambda_{tgt}$ and $\lambda_{qta}$ are hyperparameters used to balance these losses.

\section{Experiments}

\subsection{Experiment Settings}

In our study, we utilize the pre-trained Sparse 3D U-Net~\cite{graham20183d} to extract point-wise features from the 3D point clouds and the pre-trained RoBERTa~\cite{liu2019roberta} as our text encoder. These pre-trained models serve as foundational components within our architecture. The remainder of the network is trained from scratch, starting with an initial learning rate of 0.0001. To optimize this rate over the training period, we implement the PolyRL strategy, adjusting the learning rate with a decay power of $4.0$. 
Our training procedure accommodates a batch size of 32 and processes text inputs with a maximum sentence length of 80 characters. To segment the original point cloud into manageable units, we adopt an unsupervised method for generating superpoints as outlined in~\cite{landrieu2018large,sun2023superpoint}. In TSQ, we first select a total of 256 seed queries using Farthest Superpoint Sampling. Subsequently, after filtering based on text correlation scores $R$, we end up with a final set of 128 queries, designated as $N_Q$.
The MDIN is configured with six stacked layers to ensure robust feature integration and segmentation performance. We finely tune our network with specific loss weight settings: $\lambda_{qgd}$ is adjusted to $5$, ensuring that query guidance loss is emphasized, while $\lambda_{mask}$ remains at $1$. The weights for target and query-target alignment losses, $\lambda_{tgt}$ and $\lambda_{qta}$, are set more conservatively at $0.1$ to balance the training dynamics. 
All experiments are conducted using the PyTorch framework on a single NVIDIA GeForce RTX 3090 GPU.

\subsection{Quantitative Comparison}

\noindent$\bullet$~\textbf{3D-GRES Results.} We present results of 3D-REC and RES models on the 3D-GRES task in Tab.~\ref{3dgres}, trained on the Multi3DRes benchmark. For single-stage networks like 3D-STMN~\cite{wu20233d}, zero-targets are predicted for samples with fewer than 50 positive points~\cite{liu2023gres}. Two-stage networks follow M3DRef-CLIP's~\cite{zhang2023multi3drefer} approach, classifying zero-targets if all instance scores are below $\tau_{pred}$.

Using ReLA~\cite{liu2023gres} from 2D-GRES as our 3D baseline, we adapt it for 3D by initializing learnable queries similar to DETR~\cite{carion2020detr}, achieving an mIoU of 42.8\%. M3DRef-CLIP performs slightly lower, and 3D-STMN, despite being state-of-the-art for 3D-RES, struggles with complex multi-object scenarios.

Our proposed model achieves the highest mIoU of 47.5\%, surpassing M3DRef-CLIP by 10.1 points, and improves multiple target Acc@0.25 by 9.5 points, demonstrating strong discriminative capabilities for 3D-GRES.

\noindent$\bullet$~\textbf{Traditional 3D-RES Results.} As shown in Tab.~\ref{tab:scanrefer}, our proposed MDIN achieves substantial improvements over the previous state-of-the-art 3D-STMN~\cite{wu20233d} in traditional 3D-RES tasks. MDIN exhibits a 13.3-point increase in Acc@0.5 and an 8.8-point increase in mIoU. Notably, in scenes with multiple disruptive instances ("multiple"), MDIN demonstrates even greater enhancements, with a 10.3-point boost in mIoU and a 15.7-point increase in Acc@0.5. This indicates that our query decoupling approach not only improves traditional single-target 3D-RES tasks but also enhances semantic understanding and instance-level reasoning.

\begin{table}
    \centering
    \caption{The traditional 3D-RES results on ScanRefer. \dag~The mIoU and accuracy are reevaluated on our machine.}
    \resizebox{1\linewidth}{!}{
        \begin{tabular}{c|c|cc|cc|ccc}
            \toprule
            & & \multicolumn{2}{c|}{\textbf{Unique} ($\sim$19\%)} 
            & \multicolumn{2}{c|}{\textbf{Multiple} ($\sim$81\%)} 
            & \multicolumn{3}{c}{\textbf{Overall}} \\
            
            \multirow{-2}{*}{\centering \textbf{Method}}
            & \multirow{-2}{*}{\textbf{Reference}} 
            & 0.25 & 0.5 
            & 0.25 & 0.5 
            & \textbf{0.25} & \textbf{0.5} & \textbf{mIoU} \\
            \midrule

            TGNN~\cite{huang2021text} 
            & AAAI'21
            & - & - 
            & - & - 
            & 37.5 & 31.4 & 27.8 \\  

            TGNN\dag~\cite{huang2021text} 
            & AAAI'21
            & {69.3} & {57.8} 
            & {31.2} & {26.6} 
            & {38.6} & {32.7} & {28.8} \\   

            InstanceRefer\dag~\cite{yuan2021instancerefer} 
            & ICCV'21
            & 81.6 & 72.2 
            & 29.4 & 23.5 
            & 40.2 & 33.5 & 30.6  \\
            
            X-RefSeg3D~\cite{qian2024x}  
            & AAAI'24
            & {-} & {-} 
            & {-} & {-} 
            & {40.3} & {33.8} & {29.9}  \\

            3D-STMN~\cite{wu20233d}  
            & AAAI'24
            & {89.3} & {84.0} 
            & {46.2} & {29.2} 
            & {54.6} & {39.8} & {39.5}  \\ 

            SegPoint~\cite{he2024segpoint}
            & ACMMM'24
            & {-} & {-} 
            & {-} & {-} 
            & - & - & 41.7 \\
            
            RefMask3D~\cite{he2024refmask3d}
            & ACMMM'24
            & 89.6 & 84.7 
            & 48.1 & 40.8 
            & 55.9 & 49.2 & 44.9 \\ 

            \textbf{MDIN (Ours)}
            & ACMMM'24
            & \textbf{91.0} & \textbf{87.2} 
            & \textbf{50.1} & \textbf{44.9} 
            & \textbf{58.0} & \textbf{53.1} & \textbf{48.3}  \\
            \bottomrule
        \end{tabular}
    }
    \label{tab:scanrefer}
\end{table}

\subsection{Ablation Study}

\subsubsection{Ablation Study on Decoupling Modeling}

The decoupling in MDIN is achieved through TSQ and MDO. TSQ decouples queries in the physical space initially, while MDO refines them in the semantic space during optimization. We conducted an ablation study to assess their impact on 3D-GRES performance, as detailed in Tab.~\ref{tab2}. The results indicate a significant performance drop when neither module is used. Incorporating TSQ improves the model's IoU by 1.6 points by refining queries and removing redundancy. Adding MDO further enhances performance, yielding a 4.5-point improvement in the challenging multiple-target Acc@0.5 metric, due to better query decoupling and semantic consistency.

\subsubsection{Ablation Study on Losses}

We conducted ablation studies on the loss components, as shown in Table 4. Rows 1 and 2 demonstrate that using $\mathcal{L}_{tgt}$ for query decoupling significantly boosts performance, with a 3.6-point increase in Acc@0.5 for zero targets with distractors. The inclusion of TSQ's $\mathcal{L}_{qgd}$, as seen in rows 2 and 3, and rows 4 and 5, substantially improves the Acc@0.25 metric for single targets, emphasizing the benefit of removing text-irrelevant redundancy for enhanced discriminability. Finally, $\mathcal{L}_{qta}$ notably increases the Acc@0.5 for multiple targets by 3 points (rows 3 and 5), due to its role in maintaining semantic consistency and selectively filtering queries aligned with specific semantic details.

\begin{table}
\renewcommand{\arraystretch}{1.2}
\centering
\caption{Ablation studies on designed modules, where ``zt w/ dis'' means zero target with distractor, ``st w/ dis'' means single target with distractor, ``mt'' means multiple target.}
\resizebox{1.0\linewidth}{!}{
\begin{tabular}{lcc|c|c|cccc} 
\toprule
 & \multirow{2}{*}{\centering \textbf{TSQ}} &
   \multirow{2}{*}{\centering \textbf{MOD}} & 
   \multirow{2}{*}{\centering \textbf{mIoU}} &
   \textbf{Acc@0.25} & 
   \multicolumn{4}{c}{\textbf{Acc@0.5}}  \\
  &  &  &  & \textbf{Overall} &\textbf{zt w/ dis}  &\textbf{st w/ dis}  &\textbf{mt}  &\textbf{Overall}\\
\midrule
1 & $\times$ & $\times$  & 42.8 & 61.8 & 36.2 & 20.4 & 42.4 & 37.4 \\
2 & \checkmark & $\times$ & 44.4 & 63.2  & 42.5 & 24.6 & 42.3 & 40.5\\
3 & \checkmark & \checkmark & \textbf{47.5} & \textbf{67.0} & \textbf{47.9} & \textbf{29.5} & \textbf{46.8} & \textbf{44.7} \\
\bottomrule
\end{tabular}
}
\label{tab2}
\end{table}

\begin{table}
\renewcommand{\arraystretch}{1.2}
\centering
\caption{Ablation studies on the components of the loss.}
\resizebox{1.0\linewidth}{!}{
\begin{tabular}{lccc|c|c|cccc} 
\toprule
 & \multirow{2}{*}{\centering $\mathcal{L}_{qgd}$} &
   \multirow{2}{*}{\centering $\mathcal{L}_{tgt}$} & 
   \multirow{2}{*}{\centering $\mathcal{L}_{qta}$} &
   \multirow{2}{*}{\centering \textbf{mIoU}} & \textbf{Acc@0.25} & 
   \multicolumn{4}{c}{\textbf{Acc@0.5}}  \\
  &  &  &  &  & \textbf{Overall} &\textbf{zt w/ dis}  &\textbf{st w/ dis}  &\textbf{mt}  &\textbf{Overall}\\
\midrule
1 & $\times$ & $\times$ & $\times$ & 43.3 & 61.4 & 42.2  & 22.3 & 41.1 & 38.8 \\
2 & $\times$ & \checkmark & $\times$ & 45.2 & 65.5 & 45.8 & 24.6 & 43.1 & 41.1 \\
3 & \checkmark & \checkmark & $\times$ & 46.4 & 65.5 & 46.8 & 26.3 & 43.8 & 42.4\\
4 & $\times$ & \checkmark & \checkmark & 46.2 & 65.9 & 46.5 & 26.2 & 45.3 & 42.3\\
5 & \checkmark & \checkmark & \checkmark & \textbf{47.5} & \textbf{67.0} & \textbf{47.9} & \textbf{29.5} & \textbf{46.8} & \textbf{44.7} \\
\bottomrule
\end{tabular}
}
\label{tab4}
\end{table}

\begin{figure}
    \centering
    \includegraphics[width=0.47\textwidth]{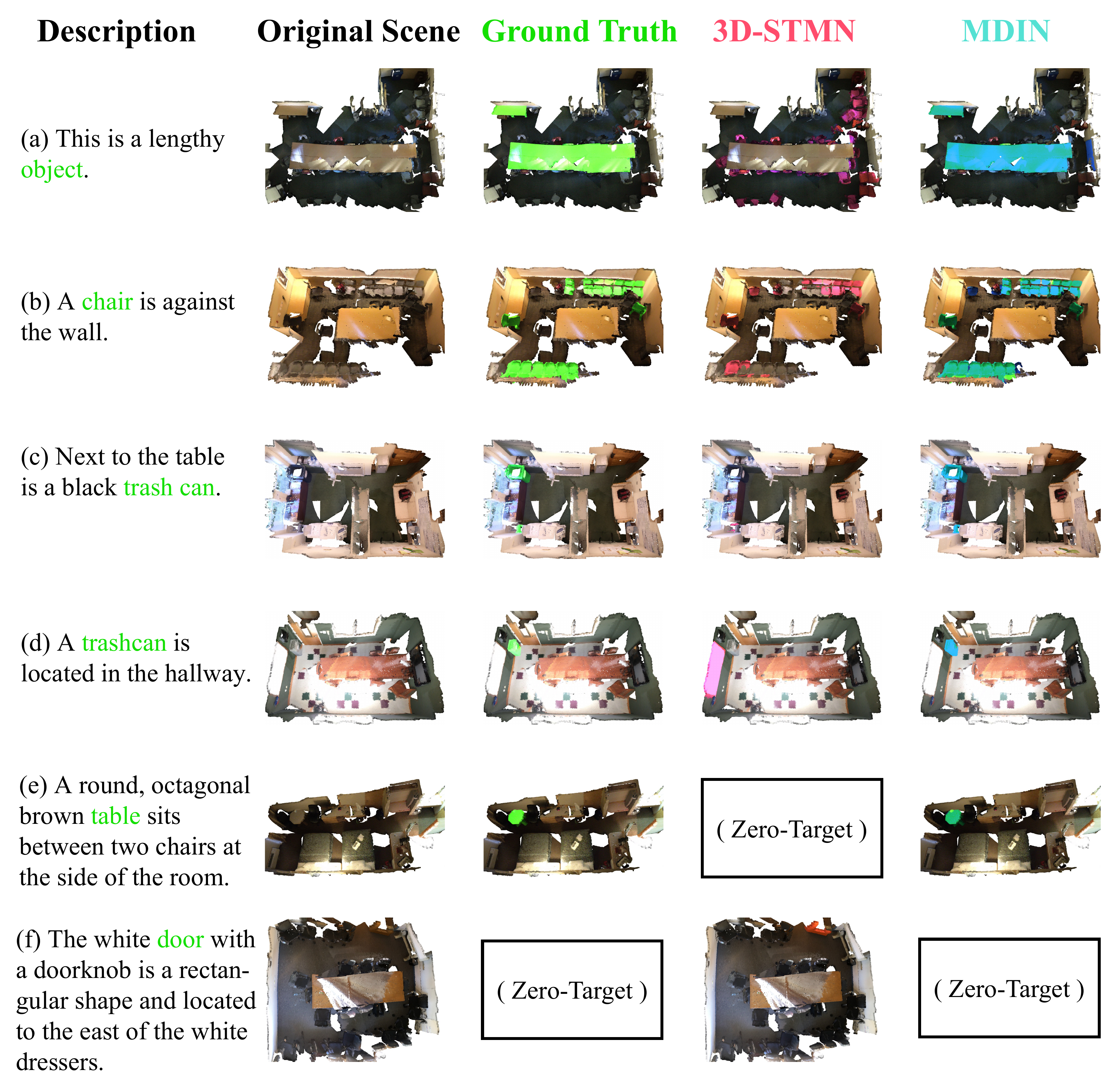}
    \caption{Qualitative comparison between the proposed MDIN and 3D-STMN. Zoom in for the best view.}
    \label{fig:3}
\end{figure}

\section{Visualization}

We visually compared the competitive 3D-STMN and MDIN models on the Multi3DRefer validation set, as shown in Fig.~\ref{fig:3}. MDIN excels in accurately segmenting challenging single-target cases and shows strong discriminative abilities for multiple and zero-target scenarios.

In instances with multiple objects, such as \textbf{(a)}, \textbf{(b)}, and \textbf{(c)} in Fig.~\ref{fig:3}, 3D-STMN struggles without decoupling, leading to semantic misidentification or missing targets. Conversely, MDIN effectively understands target semantics and segments all relevant objects. MDIN also performs well on single-target cases, as demonstrated in \textbf{(d)} and \textbf{(e)}. For extremely challenging non-target samples, like \textbf{(f)}, 3D-STMN’s reliance on semantic feature similarity results in errors, while MDIN’s decoupled approach correctly identifies and excludes non-target objects, making accurate zero-target predictions.

\section{Conclusion}

In this paper, we introduce 3D-GRES, a novel task for segmenting an arbitrary number of instances based on referential descriptions in 3D point clouds. We propose the Multi-Query Decoupled Interaction Network (MDIN) to address this task. MDIN features two key components: Text-driven Sparse Queries (TSQ) and Multi-object Decoupling Optimization (MDO). TSQ refines query generation by removing redundancies and retaining queries relevant to the textual context, while MDO enhances semantic consistency and query decoupling through various loss functions. Together, these components enable MDIN to set a new benchmark for performance on the 3D-GRES task, providing a robust and modular solution.

\begin{acks}
This work was supported by National Key R\&D Program of China (No. 2022ZD0118201), the National Science Fund for Distinguished Young Scholars (No. 62025603), the National Natural Science Foundation of China (No. U21B2037, No. U22B2051, No. 62072389, No. U21A20472), the National Natural Science Fund for Young Scholars of China (No. 62302411), China Postdoctoral Science Foundation (No. 2023M732948), the Natural Science Foundation of Fujian Province of China (No. 2021J01002,  No. 2022J06001), and partially sponsored by CCF-NetEase ThunderFire Innovation Research Funding (NO. CCF-Netease 202301).
\end{acks}

\bibliographystyle{ACM-Reference-Format}
\balance
\bibliography{3D-GRES}


\begin{thebibliography}{86}


\ifx \showCODEN    \undefined \def \showCODEN     #1{\unskip}     \fi
\ifx \showDOI      \undefined \def \showDOI       #1{#1}\fi
\ifx \showISBNx    \undefined \def \showISBNx     #1{\unskip}     \fi
\ifx \showISBNxiii \undefined \def \showISBNxiii  #1{\unskip}     \fi
\ifx \showISSN     \undefined \def \showISSN      #1{\unskip}     \fi
\ifx \showLCCN     \undefined \def \showLCCN      #1{\unskip}     \fi
\ifx \shownote     \undefined \def \shownote      #1{#1}          \fi
\ifx \showarticletitle \undefined \def \showarticletitle #1{#1}   \fi
\ifx \showURL      \undefined \def \showURL       {\relax}        \fi
\providecommand\bibfield[2]{#2}
\providecommand\bibinfo[2]{#2}
\providecommand\natexlab[1]{#1}
\providecommand\showeprint[2][]{arXiv:#2}

\bibitem[Achlioptas et~al\mbox{.}(2020)]%
        {achlioptas2020referit3d}
\bibfield{author}{\bibinfo{person}{Panos Achlioptas}, \bibinfo{person}{Ahmed Abdelreheem}, \bibinfo{person}{Fei Xia}, \bibinfo{person}{Mohamed Elhoseiny}, {and} \bibinfo{person}{Leonidas Guibas}.} \bibinfo{year}{2020}\natexlab{}.
\newblock \showarticletitle{Referit3d: Neural listeners for fine-grained 3d object identification in real-world scenes}. In \bibinfo{booktitle}{\emph{Computer Vision--ECCV 2020: 16th European Conference, Glasgow, UK, August 23--28, 2020, Proceedings, Part I 16}}. Springer, \bibinfo{pages}{422--440}.
\newblock


\bibitem[Carion et~al\mbox{.}(2020)]%
        {carion2020detr}
\bibfield{author}{\bibinfo{person}{Nicolas Carion}, \bibinfo{person}{Francisco Massa}, \bibinfo{person}{Gabriel Synnaeve}, \bibinfo{person}{Nicolas Usunier}, \bibinfo{person}{Alexander Kirillov}, {and} \bibinfo{person}{Sergey Zagoruyko}.} \bibinfo{year}{2020}\natexlab{}.
\newblock \showarticletitle{End-to-end object detection with transformers}. In \bibinfo{booktitle}{\emph{European conference on computer vision}}. Springer, \bibinfo{pages}{213--229}.
\newblock


\bibitem[Chen et~al\mbox{.}(2020)]%
        {chen2020scanrefer}
\bibfield{author}{\bibinfo{person}{Dave~Zhenyu Chen}, \bibinfo{person}{Angel~X Chang}, {and} \bibinfo{person}{Matthias Nie{\ss}ner}.} \bibinfo{year}{2020}\natexlab{}.
\newblock \showarticletitle{Scanrefer: 3d object localization in rgb-d scans using natural language}. In \bibinfo{booktitle}{\emph{European conference on computer vision}}. Springer, \bibinfo{pages}{202--221}.
\newblock


\bibitem[Chen et~al\mbox{.}(2022)]%
        {chen2022language}
\bibfield{author}{\bibinfo{person}{Shizhe Chen}, \bibinfo{person}{Pierre-Louis Guhur}, \bibinfo{person}{Makarand Tapaswi}, \bibinfo{person}{Cordelia Schmid}, {and} \bibinfo{person}{Ivan Laptev}.} \bibinfo{year}{2022}\natexlab{}.
\newblock \showarticletitle{Language conditioned spatial relation reasoning for 3d object grounding}.
\newblock \bibinfo{journal}{\emph{Advances in neural information processing systems}}  \bibinfo{volume}{35} (\bibinfo{year}{2022}), \bibinfo{pages}{20522--20535}.
\newblock


\bibitem[Cheng et~al\mbox{.}(2021)]%
        {cheng2021back}
\bibfield{author}{\bibinfo{person}{Bowen Cheng}, \bibinfo{person}{Lu Sheng}, \bibinfo{person}{Shaoshuai Shi}, \bibinfo{person}{Ming Yang}, {and} \bibinfo{person}{Dong Xu}.} \bibinfo{year}{2021}\natexlab{}.
\newblock \showarticletitle{Back-tracing representative points for voting-based 3d object detection in point clouds}. In \bibinfo{booktitle}{\emph{CVPR}}. \bibinfo{pages}{8963--8972}.
\newblock


\bibitem[Dai et~al\mbox{.}(2017)]%
        {dai2017scannet}
\bibfield{author}{\bibinfo{person}{Angela Dai}, \bibinfo{person}{Angel~X Chang}, \bibinfo{person}{Manolis Savva}, \bibinfo{person}{Maciej Halber}, \bibinfo{person}{Thomas Funkhouser}, {and} \bibinfo{person}{Matthias Nie{\ss}ner}.} \bibinfo{year}{2017}\natexlab{}.
\newblock \showarticletitle{Scannet: Richly-annotated 3d reconstructions of indoor scenes}. In \bibinfo{booktitle}{\emph{Proceedings of the IEEE conference on computer vision and pattern recognition}}. \bibinfo{pages}{5828--5839}.
\newblock


\bibitem[Dang et~al\mbox{.}(2023)]%
        {dang2023instructdet}
\bibfield{author}{\bibinfo{person}{Ronghao Dang}, \bibinfo{person}{Jiangyan Feng}, \bibinfo{person}{Haodong Zhang}, \bibinfo{person}{Chongjian Ge}, \bibinfo{person}{Lin Song}, \bibinfo{person}{Lijun Gong}, \bibinfo{person}{Chengju Liu}, \bibinfo{person}{Qijun Chen}, \bibinfo{person}{Feng Zhu}, \bibinfo{person}{Rui Zhao}, {et~al\mbox{.}}} \bibinfo{year}{2023}\natexlab{}.
\newblock \showarticletitle{Instructdet: Diversifying referring object detection with generalized instructions}.
\newblock \bibinfo{journal}{\emph{arXiv preprint arXiv:2310.05136}} (\bibinfo{year}{2023}).
\newblock


\bibitem[Deng et~al\mbox{.}(2018)]%
        {deng2018visual}
\bibfield{author}{\bibinfo{person}{Chaorui Deng}, \bibinfo{person}{Qi Wu}, \bibinfo{person}{Qingyao Wu}, \bibinfo{person}{Fuyuan Hu}, \bibinfo{person}{Fan Lyu}, {and} \bibinfo{person}{Mingkui Tan}.} \bibinfo{year}{2018}\natexlab{}.
\newblock \showarticletitle{Visual grounding via accumulated attention}. In \bibinfo{booktitle}{\emph{Proceedings of the IEEE conference on computer vision and pattern recognition}}. \bibinfo{pages}{7746--7755}.
\newblock


\bibitem[Devlin et~al\mbox{.}(2018)]%
        {devlin2018bert}
\bibfield{author}{\bibinfo{person}{Jacob Devlin}, \bibinfo{person}{Ming-Wei Chang}, \bibinfo{person}{Kenton Lee}, {and} \bibinfo{person}{Kristina Toutanova}.} \bibinfo{year}{2018}\natexlab{}.
\newblock \showarticletitle{Bert: Pre-training of deep bidirectional transformers for language understanding}.
\newblock \bibinfo{journal}{\emph{arXiv preprint arXiv:1810.04805}} (\bibinfo{year}{2018}).
\newblock


\bibitem[Ding et~al\mbox{.}(2021)]%
        {ding2021vision}
\bibfield{author}{\bibinfo{person}{Henghui Ding}, \bibinfo{person}{Chang Liu}, \bibinfo{person}{Suchen Wang}, {and} \bibinfo{person}{Xudong Jiang}.} \bibinfo{year}{2021}\natexlab{}.
\newblock \showarticletitle{Vision-language transformer and query generation for referring segmentation}. In \bibinfo{booktitle}{\emph{ICCV}}. \bibinfo{pages}{16321--16330}.
\newblock


\bibitem[Ding et~al\mbox{.}(2023)]%
        {VLT}
\bibfield{author}{\bibinfo{person}{Henghui Ding}, \bibinfo{person}{Chang Liu}, \bibinfo{person}{Suchen Wang}, {and} \bibinfo{person}{Xudong Jiang}.} \bibinfo{year}{2023}\natexlab{}.
\newblock \showarticletitle{{VLT}: Vision-language transformer and query generation for referring segmentation}.
\newblock \bibinfo{journal}{\emph{IEEE Transactions on Pattern Analysis and Machine Intelligence}} \bibinfo{volume}{45}, \bibinfo{number}{6} (\bibinfo{year}{2023}).
\newblock


\bibitem[Fei et~al\mbox{.}(2023)]%
        {fei2023scene}
\bibfield{author}{\bibinfo{person}{Hao Fei}, \bibinfo{person}{Qian Liu}, \bibinfo{person}{Meishan Zhang}, \bibinfo{person}{Min Zhang}, {and} \bibinfo{person}{Tat-Seng Chua}.} \bibinfo{year}{2023}\natexlab{}.
\newblock \showarticletitle{Scene Graph as Pivoting: Inference-time Image-free Unsupervised Multimodal Machine Translation with Visual Scene Hallucination}. In \bibinfo{booktitle}{\emph{Proceedings of the 61st Annual Meeting of the Association for Computational Linguistics}}. \bibinfo{pages}{5980--5994}.
\newblock


\bibitem[Fei et~al\mbox{.}(2024a)]%
        {fei2024dysen}
\bibfield{author}{\bibinfo{person}{Hao Fei}, \bibinfo{person}{Shengqiong Wu}, \bibinfo{person}{Wei Ji}, \bibinfo{person}{Hanwang Zhang}, {and} \bibinfo{person}{Tat-Seng Chua}.} \bibinfo{year}{2024}\natexlab{a}.
\newblock \showarticletitle{Dysen-VDM: Empowering Dynamics-aware Text-to-Video Diffusion with LLMs}. In \bibinfo{booktitle}{\emph{CVPR}}. \bibinfo{pages}{7641--7653}.
\newblock


\bibitem[Fei et~al\mbox{.}(2024b)]%
        {fei2024enhancing}
\bibfield{author}{\bibinfo{person}{Hao Fei}, \bibinfo{person}{Shengqiong Wu}, \bibinfo{person}{Meishan Zhang}, \bibinfo{person}{Min Zhang}, \bibinfo{person}{Tat-Seng Chua}, {and} \bibinfo{person}{Shuicheng Yan}.} \bibinfo{year}{2024}\natexlab{b}.
\newblock \showarticletitle{Enhancing video-language representations with structural spatio-temporal alignment}.
\newblock \bibinfo{journal}{\emph{IEEE Transactions on Pattern Analysis and Machine Intelligence}} (\bibinfo{year}{2024}).
\newblock


\bibitem[Feng et~al\mbox{.}(2021)]%
        {feng2021free}
\bibfield{author}{\bibinfo{person}{Mingtao Feng}, \bibinfo{person}{Zhen Li}, \bibinfo{person}{Qi Li}, \bibinfo{person}{Liang Zhang}, \bibinfo{person}{XiangDong Zhang}, \bibinfo{person}{Guangming Zhu}, \bibinfo{person}{Hui Zhang}, \bibinfo{person}{Yaonan Wang}, {and} \bibinfo{person}{Ajmal Mian}.} \bibinfo{year}{2021}\natexlab{}.
\newblock \showarticletitle{Free-form description guided 3d visual graph network for object grounding in point cloud}. In \bibinfo{booktitle}{\emph{ICCV}}. \bibinfo{pages}{3722--3731}.
\newblock


\bibitem[Graham et~al\mbox{.}(2018)]%
        {graham20183d}
\bibfield{author}{\bibinfo{person}{Benjamin Graham}, \bibinfo{person}{Martin Engelcke}, {and} \bibinfo{person}{Laurens Van Der~Maaten}.} \bibinfo{year}{2018}\natexlab{}.
\newblock \showarticletitle{3d semantic segmentation with submanifold sparse convolutional networks}. In \bibinfo{booktitle}{\emph{Proceedings of the IEEE conference on computer vision and pattern recognition}}. \bibinfo{pages}{9224--9232}.
\newblock


\bibitem[He et~al\mbox{.}(2021)]%
        {he2021transrefer3d}
\bibfield{author}{\bibinfo{person}{Dailan He}, \bibinfo{person}{Yusheng Zhao}, \bibinfo{person}{Junyu Luo}, \bibinfo{person}{Tianrui Hui}, \bibinfo{person}{Shaofei Huang}, \bibinfo{person}{Aixi Zhang}, {and} \bibinfo{person}{Si Liu}.} \bibinfo{year}{2021}\natexlab{}.
\newblock \showarticletitle{Transrefer3d: Entity-and-relation aware transformer for fine-grained 3d visual grounding}. In \bibinfo{booktitle}{\emph{Proceedings of the 29th ACM International Conference on Multimedia}}. \bibinfo{pages}{2344--2352}.
\newblock


\bibitem[He and Ding(2024)]%
        {he2024refmask3d}
\bibfield{author}{\bibinfo{person}{Shuting He} {and} \bibinfo{person}{Henghui Ding}.} \bibinfo{year}{2024}\natexlab{}.
\newblock \showarticletitle{RefMask3D: Language-Guided Transformer for 3D Referring Segmentation}.
\newblock \bibinfo{journal}{\emph{arXiv preprint arXiv:2407.18244}} (\bibinfo{year}{2024}).
\newblock


\bibitem[He et~al\mbox{.}(2024)]%
        {he2024segpoint}
\bibfield{author}{\bibinfo{person}{Shuting He}, \bibinfo{person}{Henghui Ding}, \bibinfo{person}{Xudong Jiang}, {and} \bibinfo{person}{Bihan Wen}.} \bibinfo{year}{2024}\natexlab{}.
\newblock \showarticletitle{SegPoint: Segment Any Point Cloud via Large Language Model}.
\newblock \bibinfo{journal}{\emph{arXiv preprint arXiv:2407.13761}} (\bibinfo{year}{2024}).
\newblock


\bibitem[He et~al\mbox{.}(2023)]%
        {GREC}
\bibfield{author}{\bibinfo{person}{Shuting He}, \bibinfo{person}{Henghui Ding}, \bibinfo{person}{Chang Liu}, {and} \bibinfo{person}{Xudong Jiang}.} \bibinfo{year}{2023}\natexlab{}.
\newblock \showarticletitle{{GREC}: Generalized Referring Expression Comprehension}.
\newblock \bibinfo{journal}{\emph{arXiv preprint arXiv:2308.16182}} (\bibinfo{year}{2023}).
\newblock


\bibitem[Hong et~al\mbox{.}(2019)]%
        {hong2019learning}
\bibfield{author}{\bibinfo{person}{Richang Hong}, \bibinfo{person}{Daqing Liu}, \bibinfo{person}{Xiaoyu Mo}, \bibinfo{person}{Xiangnan He}, {and} \bibinfo{person}{Hanwang Zhang}.} \bibinfo{year}{2019}\natexlab{}.
\newblock \showarticletitle{Learning to compose and reason with language tree structures for visual grounding}.
\newblock \bibinfo{journal}{\emph{IEEE transactions on pattern analysis and machine intelligence}} \bibinfo{volume}{44}, \bibinfo{number}{2} (\bibinfo{year}{2019}), \bibinfo{pages}{684--696}.
\newblock


\bibitem[Hu et~al\mbox{.}(2017)]%
        {hu2017modeling}
\bibfield{author}{\bibinfo{person}{Ronghang Hu}, \bibinfo{person}{Marcus Rohrbach}, \bibinfo{person}{Jacob Andreas}, \bibinfo{person}{Trevor Darrell}, {and} \bibinfo{person}{Kate Saenko}.} \bibinfo{year}{2017}\natexlab{}.
\newblock \showarticletitle{Modeling relationships in referential expressions with compositional modular networks}. In \bibinfo{booktitle}{\emph{Proceedings of the IEEE conference on computer vision and pattern recognition}}. \bibinfo{pages}{1115--1124}.
\newblock


\bibitem[Hu et~al\mbox{.}(2016a)]%
        {hu2016segmentation}
\bibfield{author}{\bibinfo{person}{Ronghang Hu}, \bibinfo{person}{Marcus Rohrbach}, {and} \bibinfo{person}{Trevor Darrell}.} \bibinfo{year}{2016}\natexlab{a}.
\newblock \showarticletitle{Segmentation from natural language expressions}. In \bibinfo{booktitle}{\emph{Computer Vision--ECCV 2016: 14th European Conference, Amsterdam, The Netherlands, October 11--14, 2016, Proceedings, Part I 14}}. Springer, \bibinfo{pages}{108--124}.
\newblock


\bibitem[Hu et~al\mbox{.}(2016b)]%
        {hu2016natural}
\bibfield{author}{\bibinfo{person}{Ronghang Hu}, \bibinfo{person}{Huazhe Xu}, \bibinfo{person}{Marcus Rohrbach}, \bibinfo{person}{Jiashi Feng}, \bibinfo{person}{Kate Saenko}, {and} \bibinfo{person}{Trevor Darrell}.} \bibinfo{year}{2016}\natexlab{b}.
\newblock \showarticletitle{Natural language object retrieval}. In \bibinfo{booktitle}{\emph{Proceedings of the IEEE conference on computer vision and pattern recognition}}. \bibinfo{pages}{4555--4564}.
\newblock


\bibitem[Hu et~al\mbox{.}(2023)]%
        {hu2023beyond}
\bibfield{author}{\bibinfo{person}{Yutao Hu}, \bibinfo{person}{Qixiong Wang}, \bibinfo{person}{Wenqi Shao}, \bibinfo{person}{Enze Xie}, \bibinfo{person}{Zhenguo Li}, \bibinfo{person}{Jungong Han}, {and} \bibinfo{person}{Ping Luo}.} \bibinfo{year}{2023}\natexlab{}.
\newblock \showarticletitle{Beyond one-to-one: Rethinking the referring image segmentation}. In \bibinfo{booktitle}{\emph{ICCV}}. \bibinfo{pages}{4067--4077}.
\newblock


\bibitem[Hu et~al\mbox{.}(2020)]%
        {hu2020bi}
\bibfield{author}{\bibinfo{person}{Zhiwei Hu}, \bibinfo{person}{Guang Feng}, \bibinfo{person}{Jiayu Sun}, \bibinfo{person}{Lihe Zhang}, {and} \bibinfo{person}{Huchuan Lu}.} \bibinfo{year}{2020}\natexlab{}.
\newblock \showarticletitle{Bi-directional relationship inferring network for referring image segmentation}. In \bibinfo{booktitle}{\emph{CVPR}}. \bibinfo{pages}{4424--4433}.
\newblock


\bibitem[Huang et~al\mbox{.}(2021)]%
        {huang2021text}
\bibfield{author}{\bibinfo{person}{Pin-Hao Huang}, \bibinfo{person}{Han-Hung Lee}, \bibinfo{person}{Hwann-Tzong Chen}, {and} \bibinfo{person}{Tyng-Luh Liu}.} \bibinfo{year}{2021}\natexlab{}.
\newblock \showarticletitle{Text-guided graph neural networks for referring 3d instance segmentation}. In \bibinfo{booktitle}{\emph{Proceedings of the AAAI Conference on Artificial Intelligence}}, Vol.~\bibinfo{volume}{35}. \bibinfo{pages}{1610--1618}.
\newblock


\bibitem[Huang et~al\mbox{.}(2023)]%
        {huang2023dense}
\bibfield{author}{\bibinfo{person}{Wencan Huang}, \bibinfo{person}{Daizong Liu}, {and} \bibinfo{person}{Wei Hu}.} \bibinfo{year}{2023}\natexlab{}.
\newblock \showarticletitle{Dense Object Grounding in 3D Scenes}. In \bibinfo{booktitle}{\emph{Proceedings of the 31st ACM International Conference on Multimedia}}. \bibinfo{pages}{5017--5026}.
\newblock


\bibitem[Jiao et~al\mbox{.}(2021)]%
        {jiao2021two}
\bibfield{author}{\bibinfo{person}{Yang Jiao}, \bibinfo{person}{Zequn Jie}, \bibinfo{person}{Weixin Luo}, \bibinfo{person}{Jingjing Chen}, \bibinfo{person}{Yu-Gang Jiang}, \bibinfo{person}{Xiaolin Wei}, {and} \bibinfo{person}{Lin Ma}.} \bibinfo{year}{2021}\natexlab{}.
\newblock \showarticletitle{Two-stage visual cues enhancement network for referring image segmentation}. In \bibinfo{booktitle}{\emph{Proceedings of the 29th ACM international conference on multimedia}}. \bibinfo{pages}{1331--1340}.
\newblock


\bibitem[Jing et~al\mbox{.}(2021)]%
        {jing2021locate}
\bibfield{author}{\bibinfo{person}{Ya Jing}, \bibinfo{person}{Tao Kong}, \bibinfo{person}{Wei Wang}, \bibinfo{person}{Liang Wang}, \bibinfo{person}{Lei Li}, {and} \bibinfo{person}{Tieniu Tan}.} \bibinfo{year}{2021}\natexlab{}.
\newblock \showarticletitle{Locate then segment: A strong pipeline for referring image segmentation}. In \bibinfo{booktitle}{\emph{CVPR}}. \bibinfo{pages}{9858--9867}.
\newblock


\bibitem[Kazemzadeh et~al\mbox{.}(2014)]%
        {2014referitgame}
\bibfield{author}{\bibinfo{person}{Sahar Kazemzadeh}, \bibinfo{person}{Vicente Ordonez}, \bibinfo{person}{Mark Matten}, {and} \bibinfo{person}{Tamara Berg}.} \bibinfo{year}{2014}\natexlab{}.
\newblock \showarticletitle{Referitgame: Referring to objects in photographs of natural scenes}. In \bibinfo{booktitle}{\emph{Proceedings of the 2014 conference on empirical methods in natural language processing (EMNLP)}}. \bibinfo{pages}{787--798}.
\newblock


\bibitem[Lai et~al\mbox{.}(2023)]%
        {lai2023mask}
\bibfield{author}{\bibinfo{person}{Xin Lai}, \bibinfo{person}{Yuhui Yuan}, \bibinfo{person}{Ruihang Chu}, \bibinfo{person}{Yukang Chen}, \bibinfo{person}{Han Hu}, {and} \bibinfo{person}{Jiaya Jia}.} \bibinfo{year}{2023}\natexlab{}.
\newblock \showarticletitle{Mask-attention-free transformer for 3d instance segmentation}. In \bibinfo{booktitle}{\emph{ICCV}}. \bibinfo{pages}{3693--3703}.
\newblock


\bibitem[Landrieu and Simonovsky(2018)]%
        {landrieu2018large}
\bibfield{author}{\bibinfo{person}{Loic Landrieu} {and} \bibinfo{person}{Martin Simonovsky}.} \bibinfo{year}{2018}\natexlab{}.
\newblock \showarticletitle{Large-scale point cloud semantic segmentation with superpoint graphs}. In \bibinfo{booktitle}{\emph{Proceedings of the IEEE conference on computer vision and pattern recognition}}. \bibinfo{pages}{4558--4567}.
\newblock


\bibitem[Li et~al\mbox{.}(2018)]%
        {li2018referring}
\bibfield{author}{\bibinfo{person}{Ruiyu Li}, \bibinfo{person}{Kaican Li}, \bibinfo{person}{Yi-Chun Kuo}, \bibinfo{person}{Michelle Shu}, \bibinfo{person}{Xiaojuan Qi}, \bibinfo{person}{Xiaoyong Shen}, {and} \bibinfo{person}{Jiaya Jia}.} \bibinfo{year}{2018}\natexlab{}.
\newblock \showarticletitle{Referring image segmentation via recurrent refinement networks}. In \bibinfo{booktitle}{\emph{Proceedings of the IEEE Conference on Computer Vision and Pattern Recognition}}. \bibinfo{pages}{5745--5753}.
\newblock


\bibitem[Lin et~al\mbox{.}(2023)]%
        {lin2023unified}
\bibfield{author}{\bibinfo{person}{Haojia Lin}, \bibinfo{person}{Yongdong Luo}, \bibinfo{person}{Xiawu Zheng}, \bibinfo{person}{Lijiang Li}, \bibinfo{person}{Fei Chao}, \bibinfo{person}{Taisong Jin}, \bibinfo{person}{Donghao Luo}, \bibinfo{person}{Chengjie Wang}, \bibinfo{person}{Yan Wang}, {and} \bibinfo{person}{Liujuan Cao}.} \bibinfo{year}{2023}\natexlab{}.
\newblock \bibinfo{title}{A Unified Framework for 3D Point Cloud Visual Grounding}.
\newblock
\newblock
\showeprint[arxiv]{2308.11887}~[cs.CV]


\bibitem[Liu et~al\mbox{.}(2023a)]%
        {liu2023gres}
\bibfield{author}{\bibinfo{person}{Chang Liu}, \bibinfo{person}{Henghui Ding}, {and} \bibinfo{person}{Xudong Jiang}.} \bibinfo{year}{2023}\natexlab{a}.
\newblock \showarticletitle{Gres: Generalized referring expression segmentation}. In \bibinfo{booktitle}{\emph{CVPR}}. \bibinfo{pages}{23592--23601}.
\newblock


\bibitem[Liu et~al\mbox{.}(2023b)]%
        {liu2023multi}
\bibfield{author}{\bibinfo{person}{Chang Liu}, \bibinfo{person}{Henghui Ding}, \bibinfo{person}{Yulun Zhang}, {and} \bibinfo{person}{Xudong Jiang}.} \bibinfo{year}{2023}\natexlab{b}.
\newblock \showarticletitle{Multi-modal mutual attention and iterative interaction for referring image segmentation}.
\newblock \bibinfo{journal}{\emph{IEEE Transactions on Image Processing}} (\bibinfo{year}{2023}).
\newblock


\bibitem[Liu et~al\mbox{.}(2022)]%
        {liu2022instance}
\bibfield{author}{\bibinfo{person}{Chang Liu}, \bibinfo{person}{Xudong Jiang}, {and} \bibinfo{person}{Henghui Ding}.} \bibinfo{year}{2022}\natexlab{}.
\newblock \showarticletitle{Instance-specific feature propagation for referring segmentation}.
\newblock \bibinfo{journal}{\emph{IEEE Transactions on Multimedia}} (\bibinfo{year}{2022}).
\newblock


\bibitem[Liu et~al\mbox{.}(2019c)]%
        {liu2019learning}
\bibfield{author}{\bibinfo{person}{Daqing Liu}, \bibinfo{person}{Hanwang Zhang}, \bibinfo{person}{Feng Wu}, {and} \bibinfo{person}{Zheng-Jun Zha}.} \bibinfo{year}{2019}\natexlab{c}.
\newblock \showarticletitle{Learning to assemble neural module tree networks for visual grounding}. In \bibinfo{booktitle}{\emph{ICCV}}. \bibinfo{pages}{4673--4682}.
\newblock


\bibitem[Liu et~al\mbox{.}(2024)]%
        {liu2024remoteclip}
\bibfield{author}{\bibinfo{person}{Fan Liu}, \bibinfo{person}{Delong Chen}, \bibinfo{person}{Zhangqingyun Guan}, \bibinfo{person}{Xiaocong Zhou}, \bibinfo{person}{Jiale Zhu}, \bibinfo{person}{Qiaolin Ye}, \bibinfo{person}{Liyong Fu}, {and} \bibinfo{person}{Jun Zhou}.} \bibinfo{year}{2024}\natexlab{}.
\newblock \showarticletitle{Remoteclip: A vision language foundation model for remote sensing}.
\newblock \bibinfo{journal}{\emph{IEEE Transactions on Geoscience and Remote Sensing}} (\bibinfo{year}{2024}).
\newblock


\bibitem[Liu et~al\mbox{.}(2023c)]%
        {liu2023caris}
\bibfield{author}{\bibinfo{person}{Sun-Ao Liu}, \bibinfo{person}{Yiheng Zhang}, \bibinfo{person}{Zhaofan Qiu}, \bibinfo{person}{Hongtao Xie}, \bibinfo{person}{Yongdong Zhang}, {and} \bibinfo{person}{Ting Yao}.} \bibinfo{year}{2023}\natexlab{c}.
\newblock \showarticletitle{CARIS: Context-aware referring image segmentation}. In \bibinfo{booktitle}{\emph{Proceedings of the 31st ACM International Conference on Multimedia}}. \bibinfo{pages}{779--788}.
\newblock


\bibitem[Liu et~al\mbox{.}(2019b)]%
        {liu2019improving}
\bibfield{author}{\bibinfo{person}{Xihui Liu}, \bibinfo{person}{Zihao Wang}, \bibinfo{person}{Jing Shao}, \bibinfo{person}{Xiaogang Wang}, {and} \bibinfo{person}{Hongsheng Li}.} \bibinfo{year}{2019}\natexlab{b}.
\newblock \showarticletitle{Improving referring expression grounding with cross-modal attention-guided erasing}. In \bibinfo{booktitle}{\emph{CVPR}}. \bibinfo{pages}{1950--1959}.
\newblock


\bibitem[Liu et~al\mbox{.}(2019a)]%
        {liu2019roberta}
\bibfield{author}{\bibinfo{person}{Yinhan Liu}, \bibinfo{person}{Myle Ott}, \bibinfo{person}{Naman Goyal}, \bibinfo{person}{Jingfei Du}, \bibinfo{person}{Mandar Joshi}, \bibinfo{person}{Danqi Chen}, \bibinfo{person}{Omer Levy}, \bibinfo{person}{Mike Lewis}, \bibinfo{person}{Luke Zettlemoyer}, {and} \bibinfo{person}{Veselin Stoyanov}.} \bibinfo{year}{2019}\natexlab{a}.
\newblock \showarticletitle{Roberta: A robustly optimized bert pretraining approach}.
\newblock \bibinfo{journal}{\emph{arXiv preprint arXiv:1907.11692}} (\bibinfo{year}{2019}).
\newblock


\bibitem[Liu et~al\mbox{.}(2021)]%
        {liu2021group}
\bibfield{author}{\bibinfo{person}{Ze Liu}, \bibinfo{person}{Zheng Zhang}, \bibinfo{person}{Yue Cao}, \bibinfo{person}{Han Hu}, {and} \bibinfo{person}{Xin Tong}.} \bibinfo{year}{2021}\natexlab{}.
\newblock \showarticletitle{Group-free 3d object detection via transformers}. In \bibinfo{booktitle}{\emph{ICCV}}. \bibinfo{pages}{2949--2958}.
\newblock


\bibitem[Luo et~al\mbox{.}(2020a)]%
        {luo2020cascade}
\bibfield{author}{\bibinfo{person}{Gen Luo}, \bibinfo{person}{Yiyi Zhou}, \bibinfo{person}{Rongrong Ji}, \bibinfo{person}{Xiaoshuai Sun}, \bibinfo{person}{Jinsong Su}, \bibinfo{person}{Chia-Wen Lin}, {and} \bibinfo{person}{Qi Tian}.} \bibinfo{year}{2020}\natexlab{a}.
\newblock \showarticletitle{Cascade grouped attention network for referring expression segmentation}. In \bibinfo{booktitle}{\emph{Proceedings of the 28th ACM International Conference on Multimedia}}. \bibinfo{pages}{1274--1282}.
\newblock


\bibitem[Luo et~al\mbox{.}(2020b)]%
        {luo2020multi}
\bibfield{author}{\bibinfo{person}{Gen Luo}, \bibinfo{person}{Yiyi Zhou}, \bibinfo{person}{Xiaoshuai Sun}, \bibinfo{person}{Liujuan Cao}, \bibinfo{person}{Chenglin Wu}, \bibinfo{person}{Cheng Deng}, {and} \bibinfo{person}{Rongrong Ji}.} \bibinfo{year}{2020}\natexlab{b}.
\newblock \showarticletitle{Multi-task collaborative network for joint referring expression comprehension and segmentation}. In \bibinfo{booktitle}{\emph{CVPR}}. \bibinfo{pages}{10034--10043}.
\newblock


\bibitem[Luo et~al\mbox{.}(2022)]%
        {luo20223d}
\bibfield{author}{\bibinfo{person}{Junyu Luo}, \bibinfo{person}{Jiahui Fu}, \bibinfo{person}{Xianghao Kong}, \bibinfo{person}{Chen Gao}, \bibinfo{person}{Haibing Ren}, \bibinfo{person}{Hao Shen}, \bibinfo{person}{Huaxia Xia}, {and} \bibinfo{person}{Si Liu}.} \bibinfo{year}{2022}\natexlab{}.
\newblock \showarticletitle{3d-sps: Single-stage 3d visual grounding via referred point progressive selection}. In \bibinfo{booktitle}{\emph{CVPR}}. \bibinfo{pages}{16454--16463}.
\newblock


\bibitem[Ma et~al\mbox{.}(2023a)]%
        {ma2023towards}
\bibfield{author}{\bibinfo{person}{Yiwei Ma}, \bibinfo{person}{Jiayi Ji}, \bibinfo{person}{Xiaoshuai Sun}, \bibinfo{person}{Yiyi Zhou}, {and} \bibinfo{person}{Rongrong Ji}.} \bibinfo{year}{2023}\natexlab{a}.
\newblock \showarticletitle{Towards local visual modeling for image captioning}.
\newblock \bibinfo{journal}{\emph{Pattern Recognition}}  \bibinfo{volume}{138} (\bibinfo{year}{2023}), \bibinfo{pages}{109420}.
\newblock


\bibitem[Ma et~al\mbox{.}(2022)]%
        {ma2022xclip}
\bibfield{author}{\bibinfo{person}{Yiwei Ma}, \bibinfo{person}{Guohai Xu}, \bibinfo{person}{Xiaoshuai Sun}, \bibinfo{person}{Ming Yan}, \bibinfo{person}{Ji Zhang}, {and} \bibinfo{person}{Rongrong Ji}.} \bibinfo{year}{2022}\natexlab{}.
\newblock \showarticletitle{X-clip: End-to-end multi-grained contrastive learning for video-text retrieval}. In \bibinfo{booktitle}{\emph{Proceedings of the 30th ACM International Conference on Multimedia}}. \bibinfo{pages}{638--647}.
\newblock


\bibitem[Ma et~al\mbox{.}(2023b)]%
        {ma2023xmesh}
\bibfield{author}{\bibinfo{person}{Yiwei Ma}, \bibinfo{person}{Xiaoqing Zhang}, \bibinfo{person}{Xiaoshuai Sun}, \bibinfo{person}{Jiayi Ji}, \bibinfo{person}{Haowei Wang}, \bibinfo{person}{Guannan Jiang}, \bibinfo{person}{Weilin Zhuang}, {and} \bibinfo{person}{Rongrong Ji}.} \bibinfo{year}{2023}\natexlab{b}.
\newblock \showarticletitle{X-mesh: Towards fast and accurate text-driven 3d stylization via dynamic textual guidance}. In \bibinfo{booktitle}{\emph{ICCV}}. \bibinfo{pages}{2749--2760}.
\newblock


\bibitem[Mao et~al\mbox{.}(2016)]%
        {mao2016generation}
\bibfield{author}{\bibinfo{person}{Junhua Mao}, \bibinfo{person}{Jonathan Huang}, \bibinfo{person}{Alexander Toshev}, \bibinfo{person}{Oana Camburu}, \bibinfo{person}{Alan~L Yuille}, {and} \bibinfo{person}{Kevin Murphy}.} \bibinfo{year}{2016}\natexlab{}.
\newblock \showarticletitle{Generation and comprehension of unambiguous object descriptions}. In \bibinfo{booktitle}{\emph{Proceedings of the IEEE conference on computer vision and pattern recognition}}. \bibinfo{pages}{11--20}.
\newblock


\bibitem[Milletari et~al\mbox{.}(2016)]%
        {milletari2016v}
\bibfield{author}{\bibinfo{person}{Fausto Milletari}, \bibinfo{person}{Nassir Navab}, {and} \bibinfo{person}{Seyed-Ahmad Ahmadi}.} \bibinfo{year}{2016}\natexlab{}.
\newblock \showarticletitle{V-net: Fully convolutional neural networks for volumetric medical image segmentation}. In \bibinfo{booktitle}{\emph{2016 fourth international conference on 3D vision (3DV)}}. Ieee, \bibinfo{pages}{565--571}.
\newblock


\bibitem[Moenning and Dodgson(2003)]%
        {moenning2003fast}
\bibfield{author}{\bibinfo{person}{Carsten Moenning} {and} \bibinfo{person}{Neil~A Dodgson}.} \bibinfo{year}{2003}\natexlab{}.
\newblock \bibinfo{booktitle}{\emph{Fast marching farthest point sampling}}.
\newblock \bibinfo{type}{{T}echnical {R}eport}. \bibinfo{institution}{University of Cambridge, Computer Laboratory}.
\newblock


\bibitem[Nagaraja et~al\mbox{.}(2016)]%
        {nagaraja2016modeling}
\bibfield{author}{\bibinfo{person}{Varun~K Nagaraja}, \bibinfo{person}{Vlad~I Morariu}, {and} \bibinfo{person}{Larry~S Davis}.} \bibinfo{year}{2016}\natexlab{}.
\newblock \showarticletitle{Modeling context between objects for referring expression understanding}. In \bibinfo{booktitle}{\emph{Computer Vision--ECCV 2016: 14th European Conference, Amsterdam, The Netherlands, October 11--14, 2016, Proceedings, Part IV 14}}. Springer, \bibinfo{pages}{792--807}.
\newblock


\bibitem[Qi et~al\mbox{.}(2019)]%
        {qi2019deep}
\bibfield{author}{\bibinfo{person}{Charles~R Qi}, \bibinfo{person}{Or Litany}, \bibinfo{person}{Kaiming He}, {and} \bibinfo{person}{Leonidas~J Guibas}.} \bibinfo{year}{2019}\natexlab{}.
\newblock \showarticletitle{Deep hough voting for 3d object detection in point clouds}. In \bibinfo{booktitle}{\emph{ICCV}}. \bibinfo{pages}{9277--9286}.
\newblock


\bibitem[Qian et~al\mbox{.}(2024)]%
        {qian2024x}
\bibfield{author}{\bibinfo{person}{Zhipeng Qian}, \bibinfo{person}{Yiwei Ma}, \bibinfo{person}{Jiayi Ji}, {and} \bibinfo{person}{Xiaoshuai Sun}.} \bibinfo{year}{2024}\natexlab{}.
\newblock \showarticletitle{X-RefSeg3D: Enhancing Referring 3D Instance Segmentation via Structured Cross-Modal Graph Neural Networks}. In \bibinfo{booktitle}{\emph{Proceedings of the AAAI Conference on Artificial Intelligence}}, Vol.~\bibinfo{volume}{38}. \bibinfo{pages}{4551--4559}.
\newblock


\bibitem[Sadhu et~al\mbox{.}(2019)]%
        {sadhu2019zero}
\bibfield{author}{\bibinfo{person}{Arka Sadhu}, \bibinfo{person}{Kan Chen}, {and} \bibinfo{person}{Ram Nevatia}.} \bibinfo{year}{2019}\natexlab{}.
\newblock \showarticletitle{Zero-shot grounding of objects from natural language queries}. In \bibinfo{booktitle}{\emph{ICCV}}. \bibinfo{pages}{4694--4703}.
\newblock


\bibitem[Schuster et~al\mbox{.}(2015)]%
        {schuster2015generating}
\bibfield{author}{\bibinfo{person}{Sebastian Schuster}, \bibinfo{person}{Ranjay Krishna}, \bibinfo{person}{Angel Chang}, \bibinfo{person}{Li Fei-Fei}, {and} \bibinfo{person}{Christopher~D Manning}.} \bibinfo{year}{2015}\natexlab{}.
\newblock \showarticletitle{Generating semantically precise scene graphs from textual descriptions for improved image retrieval}. In \bibinfo{booktitle}{\emph{Proceedings of the fourth workshop on vision and language}}. \bibinfo{pages}{70--80}.
\newblock


\bibitem[Shi et~al\mbox{.}(2020)]%
        {shi2020points}
\bibfield{author}{\bibinfo{person}{Shaoshuai Shi}, \bibinfo{person}{Zhe Wang}, \bibinfo{person}{Jianping Shi}, \bibinfo{person}{Xiaogang Wang}, {and} \bibinfo{person}{Hongsheng Li}.} \bibinfo{year}{2020}\natexlab{}.
\newblock \showarticletitle{From points to parts: 3d object detection from point cloud with part-aware and part-aggregation network}.
\newblock \bibinfo{journal}{\emph{IEEE transactions on pattern analysis and machine intelligence}} \bibinfo{volume}{43}, \bibinfo{number}{8} (\bibinfo{year}{2020}), \bibinfo{pages}{2647--2664}.
\newblock


\bibitem[Su et~al\mbox{.}(2023)]%
        {su2023referring}
\bibfield{author}{\bibinfo{person}{Wei Su}, \bibinfo{person}{Peihan Miao}, \bibinfo{person}{Huanzhang Dou}, \bibinfo{person}{Yongjian Fu}, {and} \bibinfo{person}{Xi Li}.} \bibinfo{year}{2023}\natexlab{}.
\newblock \showarticletitle{Referring expression comprehension using language adaptive inference}. In \bibinfo{booktitle}{\emph{Proceedings of the AAAI Conference on Artificial Intelligence}}, Vol.~\bibinfo{volume}{37}. \bibinfo{pages}{2357--2365}.
\newblock


\bibitem[Sun et~al\mbox{.}(2023)]%
        {sun2023superpoint}
\bibfield{author}{\bibinfo{person}{Jiahao Sun}, \bibinfo{person}{Chunmei Qing}, \bibinfo{person}{Junpeng Tan}, {and} \bibinfo{person}{Xiangmin Xu}.} \bibinfo{year}{2023}\natexlab{}.
\newblock \showarticletitle{Superpoint transformer for 3d scene instance segmentation}. In \bibinfo{booktitle}{\emph{Proceedings of the AAAI Conference on Artificial Intelligence}}, Vol.~\bibinfo{volume}{37}. \bibinfo{pages}{2393--2401}.
\newblock


\bibitem[Vaswani et~al\mbox{.}(2017)]%
        {vaswani2017attention}
\bibfield{author}{\bibinfo{person}{Ashish Vaswani}, \bibinfo{person}{Noam Shazeer}, \bibinfo{person}{Niki Parmar}, \bibinfo{person}{Jakob Uszkoreit}, \bibinfo{person}{Llion Jones}, \bibinfo{person}{Aidan~N Gomez}, \bibinfo{person}{{\L}ukasz Kaiser}, {and} \bibinfo{person}{Illia Polosukhin}.} \bibinfo{year}{2017}\natexlab{}.
\newblock \showarticletitle{Attention is all you need}.
\newblock \bibinfo{journal}{\emph{Advances in neural information processing systems}}  \bibinfo{volume}{30} (\bibinfo{year}{2017}).
\newblock


\bibitem[Wang et~al\mbox{.}(2019)]%
        {wang2019neighbourhood}
\bibfield{author}{\bibinfo{person}{Peng Wang}, \bibinfo{person}{Qi Wu}, \bibinfo{person}{Jiewei Cao}, \bibinfo{person}{Chunhua Shen}, \bibinfo{person}{Lianli Gao}, {and} \bibinfo{person}{Anton van~den Hengel}.} \bibinfo{year}{2019}\natexlab{}.
\newblock \showarticletitle{Neighbourhood watch: Referring expression comprehension via language-guided graph attention networks}. In \bibinfo{booktitle}{\emph{CVPR}}. \bibinfo{pages}{1960--1968}.
\newblock


\bibitem[Wang et~al\mbox{.}(2023b)]%
        {wang2023unveiling}
\bibfield{author}{\bibinfo{person}{Wenxuan Wang}, \bibinfo{person}{Tongtian Yue}, \bibinfo{person}{Yisi Zhang}, \bibinfo{person}{Longteng Guo}, \bibinfo{person}{Xingjian He}, \bibinfo{person}{Xinlong Wang}, {and} \bibinfo{person}{Jing Liu}.} \bibinfo{year}{2023}\natexlab{b}.
\newblock \showarticletitle{Unveiling Parts Beyond Objects: Towards Finer-Granularity Referring Expression Segmentation}.
\newblock \bibinfo{journal}{\emph{arXiv preprint arXiv:2312.08007}} (\bibinfo{year}{2023}).
\newblock


\bibitem[Wang et~al\mbox{.}(2023a)]%
        {wang20233drp}
\bibfield{author}{\bibinfo{person}{Zehan Wang}, \bibinfo{person}{Haifeng Huang}, \bibinfo{person}{Yang Zhao}, \bibinfo{person}{Linjun Li}, \bibinfo{person}{Xize Cheng}, \bibinfo{person}{Yichen Zhu}, \bibinfo{person}{Aoxiong Yin}, {and} \bibinfo{person}{Zhou Zhao}.} \bibinfo{year}{2023}\natexlab{a}.
\newblock \showarticletitle{3drp-net: 3d relative position-aware network for 3d visual grounding}.
\newblock \bibinfo{journal}{\emph{arXiv preprint arXiv:2307.13363}} (\bibinfo{year}{2023}).
\newblock


\bibitem[Wu et~al\mbox{.}(2023b)]%
        {wu20233d}
\bibfield{author}{\bibinfo{person}{Changli Wu}, \bibinfo{person}{Yiwei Ma}, \bibinfo{person}{Qi Chen}, \bibinfo{person}{Haowei Wang}, \bibinfo{person}{Gen Luo}, \bibinfo{person}{Jiayi Ji}, {and} \bibinfo{person}{Xiaoshuai Sun}.} \bibinfo{year}{2023}\natexlab{b}.
\newblock \showarticletitle{3D-STMN: Dependency-Driven Superpoint-Text Matching Network for End-to-End 3D Referring Expression Segmentation}.
\newblock \bibinfo{journal}{\emph{arXiv preprint arXiv:2308.16632}} (\bibinfo{year}{2023}).
\newblock


\bibitem[Wu et~al\mbox{.}(2019)]%
        {wu2019unified}
\bibfield{author}{\bibinfo{person}{Hao Wu}, \bibinfo{person}{Jiayuan Mao}, \bibinfo{person}{Yufeng Zhang}, \bibinfo{person}{Yuning Jiang}, \bibinfo{person}{Lei Li}, \bibinfo{person}{Weiwei Sun}, {and} \bibinfo{person}{Wei-Ying Ma}.} \bibinfo{year}{2019}\natexlab{}.
\newblock \showarticletitle{Unified visual-semantic embeddings: Bridging vision and language with structured meaning representations}. In \bibinfo{booktitle}{\emph{CVPR}}. \bibinfo{pages}{6609--6618}.
\newblock


\bibitem[Wu et~al\mbox{.}(2024b)]%
        {wu2024towards}
\bibfield{author}{\bibinfo{person}{Jianzong Wu}, \bibinfo{person}{Xiangtai Li}, \bibinfo{person}{Xia Li}, \bibinfo{person}{Henghui Ding}, \bibinfo{person}{Yunhai Tong}, {and} \bibinfo{person}{Dacheng Tao}.} \bibinfo{year}{2024}\natexlab{b}.
\newblock \showarticletitle{Towards robust referring image segmentation}.
\newblock \bibinfo{journal}{\emph{IEEE Transactions on Image Processing}} (\bibinfo{year}{2024}).
\newblock


\bibitem[Wu et~al\mbox{.}(2024a)]%
        {wu24next}
\bibfield{author}{\bibinfo{person}{Shengqiong Wu}, \bibinfo{person}{Hao Fei}, \bibinfo{person}{Leigang Qu}, \bibinfo{person}{Wei Ji}, {and} \bibinfo{person}{Tat-Seng Chua}.} \bibinfo{year}{2024}\natexlab{a}.
\newblock \showarticletitle{NExT-GPT: Any-to-Any Multimodal LLM}. In \bibinfo{booktitle}{\emph{Proceedings of the International Conference on Machine Learning}}.
\newblock


\bibitem[Wu et~al\mbox{.}(2023a)]%
        {wu2023eda}
\bibfield{author}{\bibinfo{person}{Yanmin Wu}, \bibinfo{person}{Xinhua Cheng}, \bibinfo{person}{Renrui Zhang}, \bibinfo{person}{Zesen Cheng}, {and} \bibinfo{person}{Jian Zhang}.} \bibinfo{year}{2023}\natexlab{a}.
\newblock \showarticletitle{Eda: Explicit text-decoupling and dense alignment for 3d visual grounding}. In \bibinfo{booktitle}{\emph{CVPR}}. \bibinfo{pages}{19231--19242}.
\newblock


\bibitem[Xia et~al\mbox{.}(2023)]%
        {xia2023gsva}
\bibfield{author}{\bibinfo{person}{Zhuofan Xia}, \bibinfo{person}{Dongchen Han}, \bibinfo{person}{Yizeng Han}, \bibinfo{person}{Xuran Pan}, \bibinfo{person}{Shiji Song}, {and} \bibinfo{person}{Gao Huang}.} \bibinfo{year}{2023}\natexlab{}.
\newblock \showarticletitle{GSVA: Generalized Segmentation via Multimodal Large Language Models}.
\newblock \bibinfo{journal}{\emph{arXiv preprint arXiv:2312.10103}} (\bibinfo{year}{2023}).
\newblock


\bibitem[Xie et~al\mbox{.}(2024)]%
        {xie2024described}
\bibfield{author}{\bibinfo{person}{Chi Xie}, \bibinfo{person}{Zhao Zhang}, \bibinfo{person}{Yixuan Wu}, \bibinfo{person}{Feng Zhu}, \bibinfo{person}{Rui Zhao}, {and} \bibinfo{person}{Shuang Liang}.} \bibinfo{year}{2024}\natexlab{}.
\newblock \showarticletitle{Described Object Detection: Liberating Object Detection with Flexible Expressions}.
\newblock \bibinfo{journal}{\emph{Advances in Neural Information Processing Systems}}  \bibinfo{volume}{36} (\bibinfo{year}{2024}).
\newblock


\bibitem[Yang et~al\mbox{.}(2021)]%
        {yang2021bottom}
\bibfield{author}{\bibinfo{person}{Sibei Yang}, \bibinfo{person}{Meng Xia}, \bibinfo{person}{Guanbin Li}, \bibinfo{person}{Hong-Yu Zhou}, {and} \bibinfo{person}{Yizhou Yu}.} \bibinfo{year}{2021}\natexlab{}.
\newblock \showarticletitle{Bottom-up shift and reasoning for referring image segmentation}. In \bibinfo{booktitle}{\emph{CVPR}}. \bibinfo{pages}{11266--11275}.
\newblock


\bibitem[Yang et~al\mbox{.}(2020)]%
        {yang2020improving}
\bibfield{author}{\bibinfo{person}{Zhengyuan Yang}, \bibinfo{person}{Tianlang Chen}, \bibinfo{person}{Liwei Wang}, {and} \bibinfo{person}{Jiebo Luo}.} \bibinfo{year}{2020}\natexlab{}.
\newblock \showarticletitle{Improving one-stage visual grounding by recursive sub-query construction}. In \bibinfo{booktitle}{\emph{Computer Vision--ECCV 2020: 16th European Conference, Glasgow, UK, August 23--28, 2020, Proceedings, Part XIV 16}}. Springer, \bibinfo{pages}{387--404}.
\newblock


\bibitem[Yang et~al\mbox{.}(2022)]%
        {yang2022lavt}
\bibfield{author}{\bibinfo{person}{Zhao Yang}, \bibinfo{person}{Jiaqi Wang}, \bibinfo{person}{Yansong Tang}, \bibinfo{person}{Kai Chen}, \bibinfo{person}{Hengshuang Zhao}, {and} \bibinfo{person}{Philip~HS Torr}.} \bibinfo{year}{2022}\natexlab{}.
\newblock \showarticletitle{Lavt: Language-aware vision transformer for referring image segmentation}. In \bibinfo{booktitle}{\emph{CVPR}}. \bibinfo{pages}{18155--18165}.
\newblock


\bibitem[Ye et~al\mbox{.}(2019)]%
        {ye2019cross}
\bibfield{author}{\bibinfo{person}{Linwei Ye}, \bibinfo{person}{Mrigank Rochan}, \bibinfo{person}{Zhi Liu}, {and} \bibinfo{person}{Yang Wang}.} \bibinfo{year}{2019}\natexlab{}.
\newblock \showarticletitle{Cross-modal self-attention network for referring image segmentation}. In \bibinfo{booktitle}{\emph{CVPR}}. \bibinfo{pages}{10502--10511}.
\newblock


\bibitem[Yu et~al\mbox{.}(2018)]%
        {yu2018mattnet}
\bibfield{author}{\bibinfo{person}{Licheng Yu}, \bibinfo{person}{Zhe Lin}, \bibinfo{person}{Xiaohui Shen}, \bibinfo{person}{Jimei Yang}, \bibinfo{person}{Xin Lu}, \bibinfo{person}{Mohit Bansal}, {and} \bibinfo{person}{Tamara~L Berg}.} \bibinfo{year}{2018}\natexlab{}.
\newblock \showarticletitle{Mattnet: Modular attention network for referring expression comprehension}. In \bibinfo{booktitle}{\emph{Proceedings of the IEEE conference on computer vision and pattern recognition}}. \bibinfo{pages}{1307--1315}.
\newblock


\bibitem[Yu et~al\mbox{.}(2016)]%
        {yu2016modeling}
\bibfield{author}{\bibinfo{person}{Licheng Yu}, \bibinfo{person}{Patrick Poirson}, \bibinfo{person}{Shan Yang}, \bibinfo{person}{Alexander~C Berg}, {and} \bibinfo{person}{Tamara~L Berg}.} \bibinfo{year}{2016}\natexlab{}.
\newblock \showarticletitle{Modeling context in referring expressions}. In \bibinfo{booktitle}{\emph{Computer Vision--ECCV 2016: 14th European Conference, Amsterdam, The Netherlands, October 11-14, 2016, Proceedings, Part II 14}}. Springer, \bibinfo{pages}{69--85}.
\newblock


\bibitem[Yu et~al\mbox{.}(2017)]%
        {yu2017joint}
\bibfield{author}{\bibinfo{person}{Licheng Yu}, \bibinfo{person}{Hao Tan}, \bibinfo{person}{Mohit Bansal}, {and} \bibinfo{person}{Tamara~L Berg}.} \bibinfo{year}{2017}\natexlab{}.
\newblock \showarticletitle{A joint speaker-listener-reinforcer model for referring expressions}. In \bibinfo{booktitle}{\emph{Proceedings of the IEEE conference on computer vision and pattern recognition}}. \bibinfo{pages}{7282--7290}.
\newblock


\bibitem[Yuan et~al\mbox{.}(2021)]%
        {yuan2021instancerefer}
\bibfield{author}{\bibinfo{person}{Zhihao Yuan}, \bibinfo{person}{Xu Yan}, \bibinfo{person}{Yinghong Liao}, \bibinfo{person}{Ruimao Zhang}, \bibinfo{person}{Sheng Wang}, \bibinfo{person}{Zhen Li}, {and} \bibinfo{person}{Shuguang Cui}.} \bibinfo{year}{2021}\natexlab{}.
\newblock \showarticletitle{Instancerefer: Cooperative holistic understanding for visual grounding on point clouds through instance multi-level contextual referring}. In \bibinfo{booktitle}{\emph{ICCV}}. \bibinfo{pages}{1791--1800}.
\newblock


\bibitem[Zhang et~al\mbox{.}(2023)]%
        {zhang2023multi3drefer}
\bibfield{author}{\bibinfo{person}{Yiming Zhang}, \bibinfo{person}{ZeMing Gong}, {and} \bibinfo{person}{Angel~X Chang}.} \bibinfo{year}{2023}\natexlab{}.
\newblock \showarticletitle{Multi3drefer: Grounding text description to multiple 3d objects}. In \bibinfo{booktitle}{\emph{ICCV}}. \bibinfo{pages}{15225--15236}.
\newblock


\bibitem[Zhang et~al\mbox{.}(2016)]%
        {ZHANG201640}
\bibfield{author}{\bibinfo{person}{Yan Zhang}, \bibinfo{person}{Feng Jiang}, \bibinfo{person}{Seungmin Rho}, \bibinfo{person}{Shaohui Liu}, \bibinfo{person}{Debin Zhao}, {and} \bibinfo{person}{Rongrong Ji}.} \bibinfo{year}{2016}\natexlab{}.
\newblock \showarticletitle{3D object retrieval with multi-feature collaboration and bipartite graph matching}.
\newblock \bibinfo{journal}{\emph{Neurocomputing}}  \bibinfo{volume}{195} (\bibinfo{year}{2016}), \bibinfo{pages}{40--49}.
\newblock


\bibitem[Zhang et~al\mbox{.}(2024)]%
        {zhang2024psalm}
\bibfield{author}{\bibinfo{person}{Zheng Zhang}, \bibinfo{person}{Yeyao Ma}, \bibinfo{person}{Enming Zhang}, {and} \bibinfo{person}{Xiang Bai}.} \bibinfo{year}{2024}\natexlab{}.
\newblock \showarticletitle{PSALM: Pixelwise SegmentAtion with Large Multi-Modal Model}.
\newblock \bibinfo{journal}{\emph{arXiv preprint arXiv:2403.14598}} (\bibinfo{year}{2024}).
\newblock


\bibitem[Zhao et~al\mbox{.}(2021)]%
        {zhao20213dvg}
\bibfield{author}{\bibinfo{person}{Lichen Zhao}, \bibinfo{person}{Daigang Cai}, \bibinfo{person}{Lu Sheng}, {and} \bibinfo{person}{Dong Xu}.} \bibinfo{year}{2021}\natexlab{}.
\newblock \showarticletitle{3DVG-Transformer: Relation modeling for visual grounding on point clouds}. In \bibinfo{booktitle}{\emph{ICCV}}. \bibinfo{pages}{2928--2937}.
\newblock


\bibitem[Zhao et~al\mbox{.}(2024)]%
        {zhao2024open}
\bibfield{author}{\bibinfo{person}{Xiangyu Zhao}, \bibinfo{person}{Yicheng Chen}, \bibinfo{person}{Shilin Xu}, \bibinfo{person}{Xiangtai Li}, \bibinfo{person}{Xinjiang Wang}, \bibinfo{person}{Yining Li}, {and} \bibinfo{person}{Haian Huang}.} \bibinfo{year}{2024}\natexlab{}.
\newblock \showarticletitle{An open and comprehensive pipeline for unified object grounding and detection}.
\newblock \bibinfo{journal}{\emph{arXiv preprint arXiv:2401.02361}} (\bibinfo{year}{2024}).
\newblock


\bibitem[Zhuang et~al\mbox{.}(2018)]%
        {zhuang2018parallel}
\bibfield{author}{\bibinfo{person}{Bohan Zhuang}, \bibinfo{person}{Qi Wu}, \bibinfo{person}{Chunhua Shen}, \bibinfo{person}{Ian Reid}, {and} \bibinfo{person}{Anton Van Den~Hengel}.} \bibinfo{year}{2018}\natexlab{}.
\newblock \showarticletitle{Parallel attention: A unified framework for visual object discovery through dialogs and queries}. In \bibinfo{booktitle}{\emph{Proceedings of the IEEE Conference on Computer Vision and Pattern Recognition}}. \bibinfo{pages}{4252--4261}.
\newblock


\end{thebibliography}

\clearpage
\appendix

\section{Statistical Analysis of $N_{seed}$ in TSQ}
To explore the optimal setting of the initial sparsity level in TSQ, denoted as the number of seed queries $N_{seed}$, we conducted a statistical analysis on the Multi3DRes dataset. To facilitate quantitative description, we introduce two concepts: coverage rate and repetition rate of seed queries.
The Coverage Rate (CR) of seed queries indicates the number $N^{c}_{ins}$ of instances containing seed queries as a percentage of the total number $N_{ins}$  of instances in the scene, which can be formulated as:
\begin{equation}
     CR =  N^{c}_{ins} / N_{ins}
\end{equation}
The Repetition Rate (RR) of seed queries indicates the percentage of seed queries that repeatedly cover the same instance with other queries, out of the total number $N^{c}_q$ of seed queries covering instances, which can be formulated as:
\begin{equation}
     RR = (N^{c}_q - N^{c}_{ins}) / N^{c}_q
\end{equation}

\begin{figure}
    \centering
    \begin{subfigure}{0.45\textwidth}
        \centering
        \includegraphics[width=\linewidth]{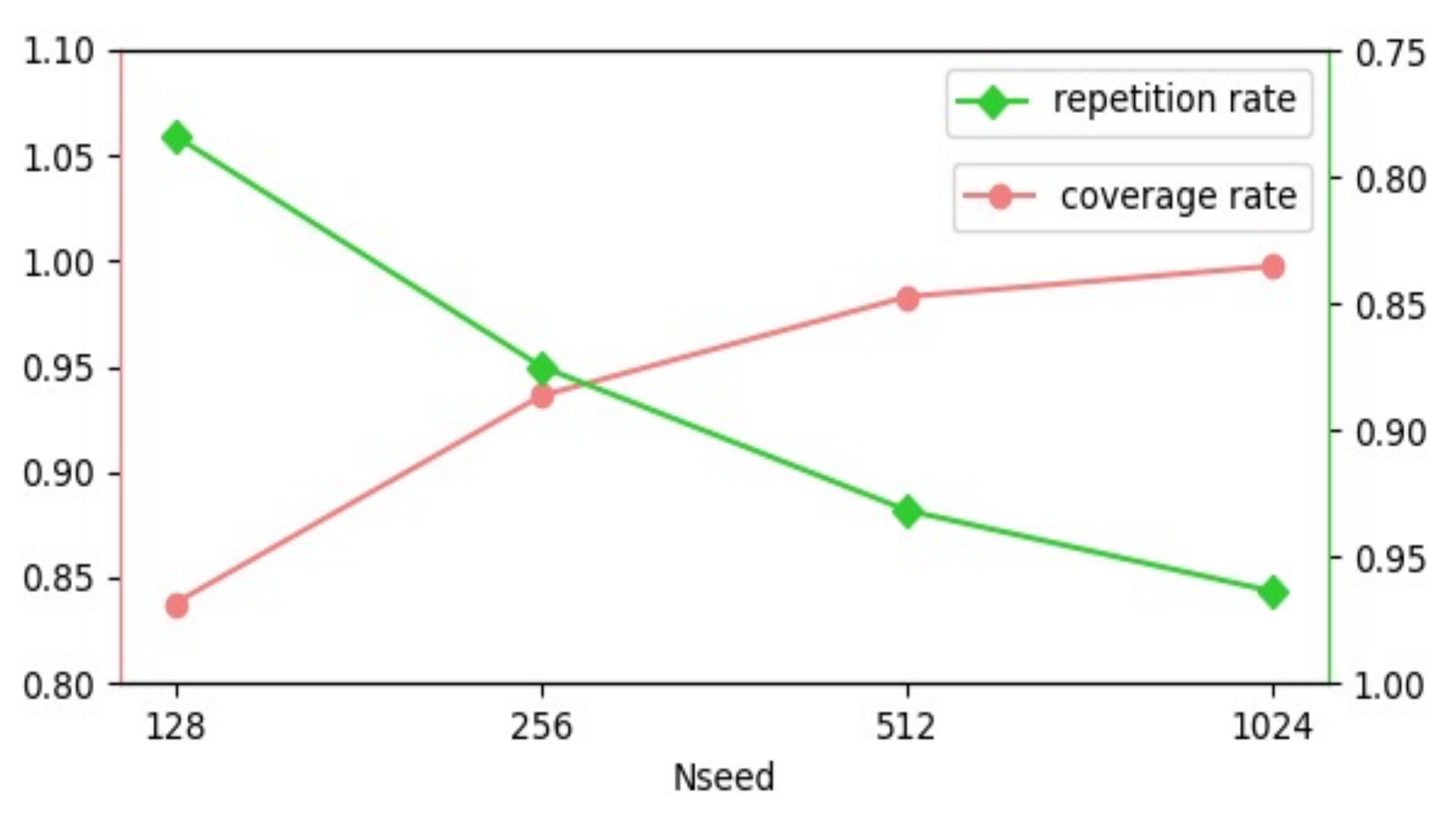}
        \caption{Instances} 
        \label{fig:a}
    \end{subfigure}
    \hfill
    \begin{subfigure}{0.45\textwidth}
        \centering
        \includegraphics[width=\linewidth]{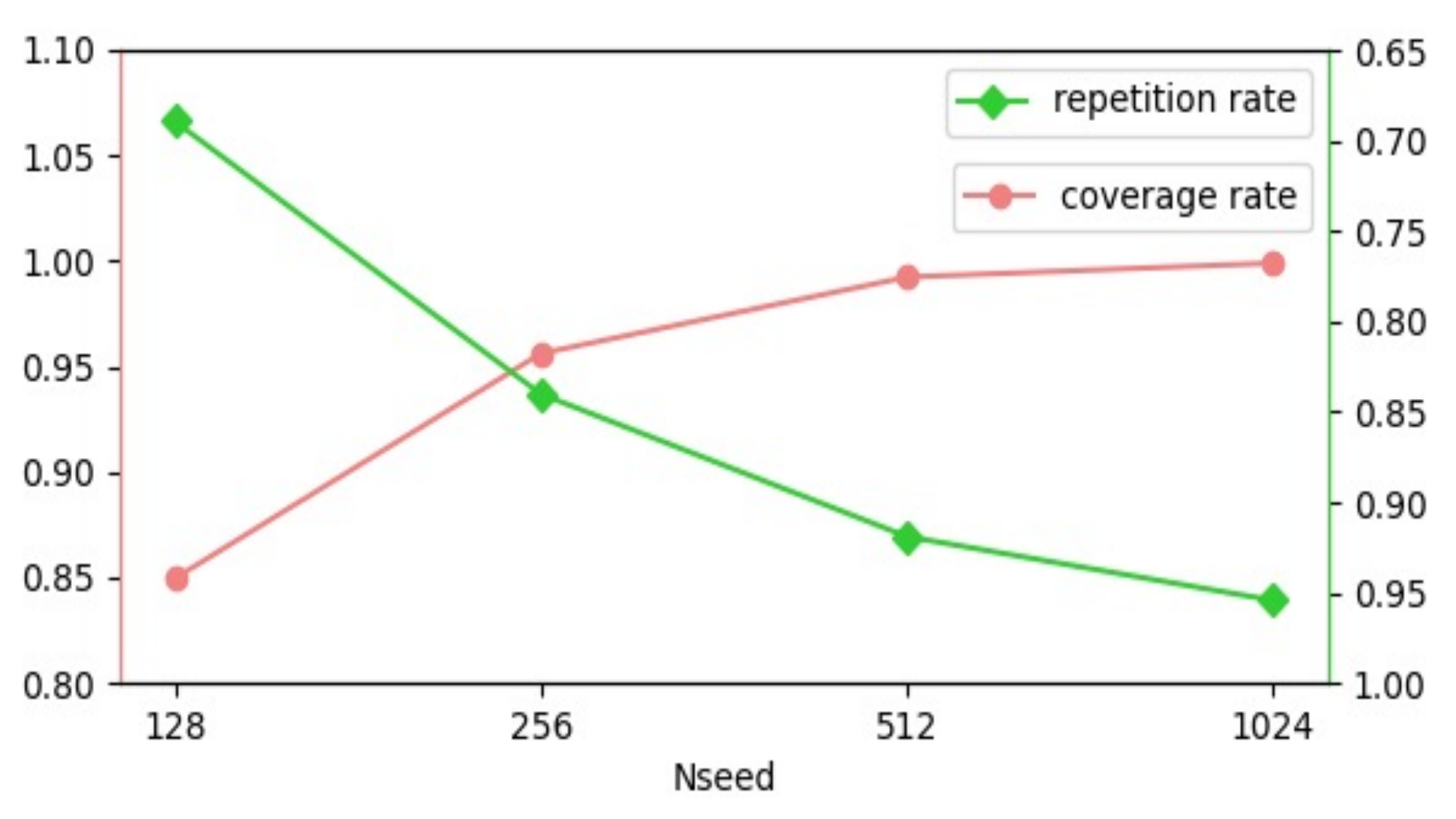}
        \caption{Ground Truth Instances}  
        \label{fig:b}
    \end{subfigure}
    \caption{The coverage rate and repetition rate of seed queries for all instances / Ground Truth instances in the scene.}
    \label{fig:ab}
\end{figure}

We conducted an analysis of the entire Multi3DRes Dataset, where we computed the coverage rate and repetition rate for different numbers $N_{seed}$ of seed queries, as shown in Fig.~\ref{fig:ab}-(a). When a large number of seed queries $N_{seed}$ are selected, it not only leads to higher coverage rate but also results in higher repetition rate. Fewer seed queries always bring a lower repetition rate, but a lower coverage rate may follow. 
We also calculated the coverage rate and repetition rate of seed queries for Ground Truth instances. As shown in Fig.~\ref{fig:ab}-(b), its trend follows the same pattern as Fig.~\ref{fig:ab}-(a). The statistical results indicate that $N_{seed} = 256$ is the optimal choice after balancing the relationship between coverage rate and repetition rate. This also validates the conclusion drawn in Sec.~5.3.2 that optimal performance is achieved when $N_{seed}$ is set to 256.

\begin{table*}
    \centering
    \caption{The results on ReferIt3D.}
    \footnotesize
    \centering
    \resizebox{1\textwidth}{!}{
    \begin{tabular}{c|ccc|ccc|ccc|ccc|ccc}
    \toprule
     & \multicolumn{3}{c|}{easy} & \multicolumn{3}{c|}{hard} & \multicolumn{3}{c|}{View Dep} & \multicolumn{3}{c|}{View Indep} & \multicolumn{3}{c}{\textbf{Overall}}  \\
     \multirow{-2}{*}{Method} & 0.25 & 0.5 & mIoU & 0.25 & 0.5 & mIoU & 0.25 & 0.5 & mIoU & 0.25 & 0.5 & mIoU & \textbf{0.25} & \textbf{0.5} & \textbf{mIoU} \\
    \hline\hline 
    \multicolumn{16}{c}{\textbf{Nr3D}}\\
    \midrule
    TGNN~\cite{huang2021text} & 29.2 & 22.3 & 21.0
    & 22.5 & 19.6 & 17.4
    & 22.2 & 18.1 & 17.0
    & 27.6 & 22.4 & 20.3
    & 25.7 & 20.9 & 19.1 \\ 
    3D-STMN~\cite{wu20233d}  & 47.9 & 31.9 & 32.6
    & 35.4 & 20.0 & 23.0
    & {37.7} & 21.1 & 24.3
    & {43.5} & 28.4 & 29.4
    & {41.5} & {25.8} & {27.6} \\ 
    \textbf{MDIN (Ours)} & \textbf{55.0} & \textbf{48.4} & \textbf{44.1}
    & \textbf{42.2} & \textbf{36.3} & \textbf{33.5}
    & \textbf{40.8} & \textbf{34.6} & \textbf{32.2}
    & \textbf{52.5} & \textbf{46.3} & \textbf{42.1}
    & \textbf{48.4} & \textbf{42.2} & \textbf{38.6} \\
    \hline\hline 
    \multicolumn{16}{c}{\textbf{Sr3D}}\\
    \midrule
    TGNN~\cite{huang2021text} & 28.2 & 23.0 & 20.9
    & 29.1 & 25.8 & 21.9
    & 23.8 & 21.3 & 18.2
    & 28.6 & 23.9 & 21.3
    & 27.5 & 22.9 & 20.2 \\
    3D-STMN~\cite{wu20233d} & 49.4 & 38.2 & 36.3 
    & 41.9 & 31.0 & 30.1 
    & {45.5} & 33.5 & 31.9
    & 47.2 & 36.2 & 34.6
    & {47.2} & {36.1} & {34.4} \\
    \textbf{MDIN (Ours)} & \textbf{58.9} & \textbf{53.2} & \textbf{48.1}
    & \textbf{51.1} & \textbf{46.8} & \textbf{42.5}
    & \textbf{53.6} & \textbf{48.7} & \textbf{44.2}
    & \textbf{56.7} & \textbf{51.4} & \textbf{46.5}
    & \textbf{56.6} & \textbf{51.3} & \textbf{46.4} \\
    \bottomrule
    \end{tabular}
    }
    \label{tab:referit3d}
\end{table*}

\section{Traditional 3D-RES}
In this section, we evaluated our MDIN on more traditional 3D-RES tasks. We compare MDIN with existing 3D-RES works on Nr3D/Sr3D datasets of ReferIt3D~\cite{achlioptas2020referit3d}. 

\subsection{Datasets}

\noindent \textbf{Nr3D and Sr3D}. Nr3D~\cite{achlioptas2020referit3d} (Natural Reference in 3D) consists of 41.5K human descriptions collected using a referring game~\cite{2014referitgame}. It describes objects in 707 ScanNet scenes. Sr3D~\cite{achlioptas2020referit3d} (Spatial Reference in 3D) contains 83.5K synthetic descriptions. It categorizes spatial relations into 5 types: horizontal proximity, vertical proximity, between, allocentric and support, and then generates descriptions using language templates.

\subsection{Quantitative Comparison on ReferIt3D}

We present the experimental results of our model on the ReferIt3D~\cite{achlioptas2020referit3d} benchmark in Tab.~\ref{tab:referit3d}. Unlike the  the original setup of ReferIt3D, we refrained from using ground truth bounding boxes or masks as input in our experiments, thereby significantly heightening the level of difficulty. Despite this heightened difficulty, our model still achieved significant improvements. Notably, MDIN achieved an Acc@0.5 gain of 16.4 points and a mIoU gain of 11.0 points on Nr3D, as well as a 15.2-point increase in Acc@0.5 and a 12.0-point increase in mIoU on Sr3D.

\section{More Ablation Studies}
\subsection{Number of  Stacked Layers in MDIN}

We investigated the impact of changing the number of stacked layers in MDIN. As shown in Tab.~\ref{tab5}, performance gradually improves with increasing layers, reaching a peak at 6 layers, followed by a slight decline.
Performance severely degrades when there is only one layer, indicating that refining layer by layer can enhance the model's reasoning ability. 
When the number of layers is excessive, gradients may become unstable, leading to a risk of training collapse.
Therefore, selecting six layers strikes a balance that yields the best model performance.

\subsection{The Textual Backbone}

In Tab.~\ref{tab6}, we compare the effects of commonly used natural language encoders.  It can be observed that that our method exhibits robustness regarding the choice of the NLP backbone. We achieve the optimal performance using Roberta~\cite{liu2019roberta}. 

\begin{table}
\renewcommand{\arraystretch}{1.4}
\centering
\caption{Ablation study on number of stacked layers.}
\resizebox{1.0\linewidth}{!}{
\begin{tabular}{cc|c|c|cccc} 
\toprule
 & \textbf{Number of}  &
   \multirow{2}{*}{\centering \textbf{mIoU}} &
   \textbf{Acc@0.25} & 
   \multicolumn{4}{c}{\textbf{Acc@0.5}}  \\
 &\textbf{Stacked Layers}  &  & \textbf{Overall} &\textbf{zt w/ dis}  &\textbf{st w/ dis}  &\textbf{mt}  &\textbf{Overall}\\
\midrule
$R1$ & 1 & 40.5 & 59.1  & 30.2 & 18.4 & 40.5 & 34.7 \\
$R2$ & 3 & 45.6 & 65.2 & 38.1 & 25.3 & 45.7 & 41.9 \\
$R3$ & 6 & \textbf{47.5} & \textbf{67.0} & \textbf{47.9} & \textbf{29.5} & \textbf{46.8} & \textbf{44.7} \\
$R4$ & 9 & 46.3 & 65.3 & 39.4 & 25.1 & 45.8 & 42.1 \\
\bottomrule
\end{tabular}
}
\label{tab5}
\end{table}

\begin{table}
\renewcommand{\arraystretch}{1.4}
\centering
\caption{Ablation study comparing text encoders.}
\resizebox{1.0\linewidth}{!}{
\begin{tabular}{cc|c|c|cccc} 
\toprule
 & \multirow{2}{*}{\centering \textbf{Text Encoder}}  &
   \multirow{2}{*}{\centering \textbf{mIoU}} &
   \textbf{Acc@0.25} & 
   \multicolumn{4}{c}{\textbf{Acc@0.5}}  \\
 &  &  & \textbf{Overall} &\textbf{zt w/ dis}  &\textbf{st w/ dis}  &\textbf{mt}  &\textbf{Overall}\\
\midrule
$R1$ & BERT-base~\cite{devlin2018bert} & 47.1 & 66.7 & 44.2 & 27.6 & 46.5 & 43.9 \\
$R2$ & BERT-large~\cite{devlin2018bert} & 47.2 & 66.9 & 44.8 & 28.1 & \textbf{46.9} &  44.1 \\
$R3$ & RoBERTa~\cite{liu2019roberta} & \textbf{47.5} & \textbf{67.0} & \textbf{47.9} & \textbf{29.5} & {46.8} & \textbf{44.7} \\
\bottomrule
\end{tabular}
}
\label{tab6}
\end{table}

\section{More Qualitative Results}

\begin{figure*}
    \centering
    \includegraphics[width=0.98\textwidth]{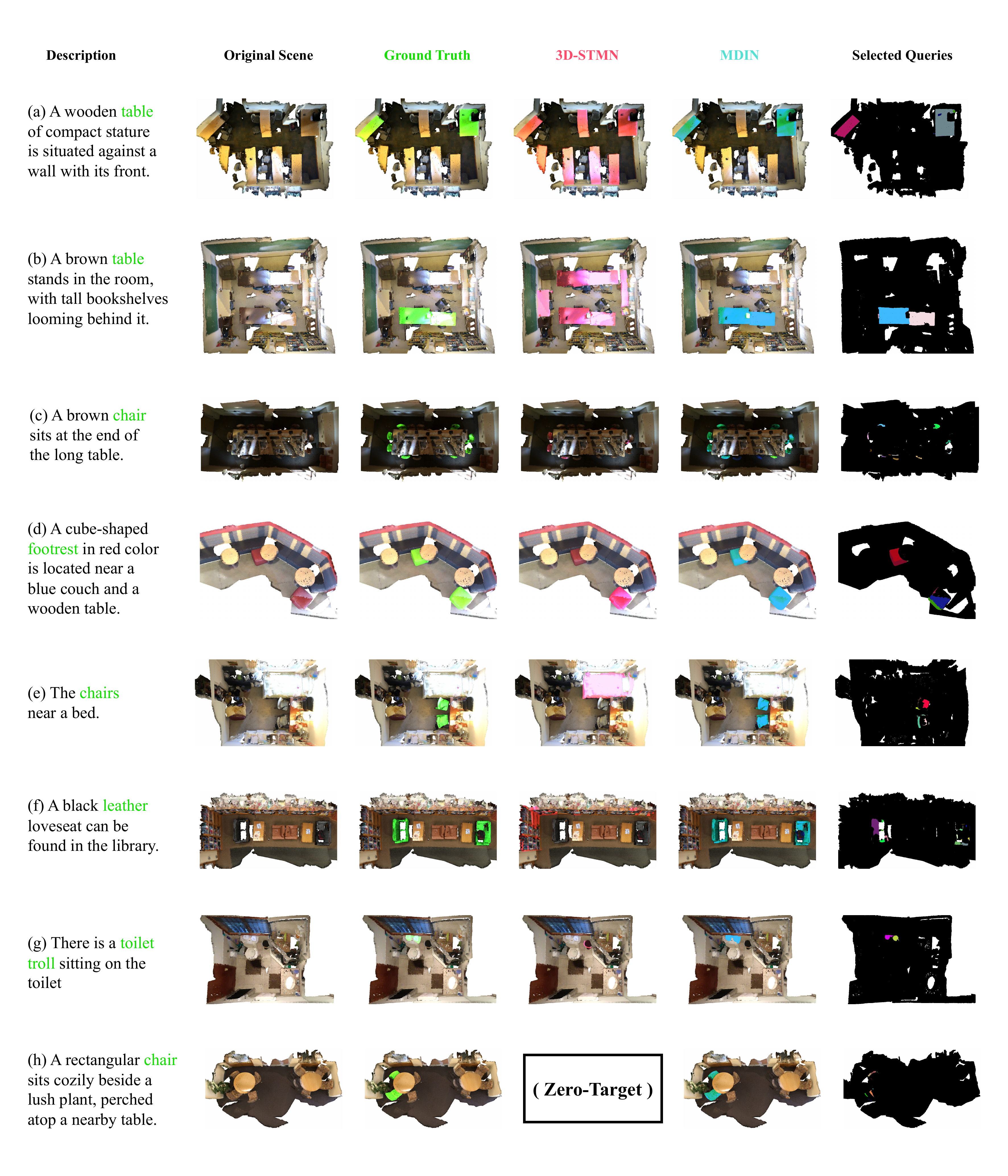}
    \caption{Qualitative comparison between the proposed MDIN and 3D-STMN on multi-targets cases. Zoom in for the best view.}
    \label{fig:6}
\end{figure*}

\begin{figure*}
    \centering
    \includegraphics[width=0.98\textwidth]{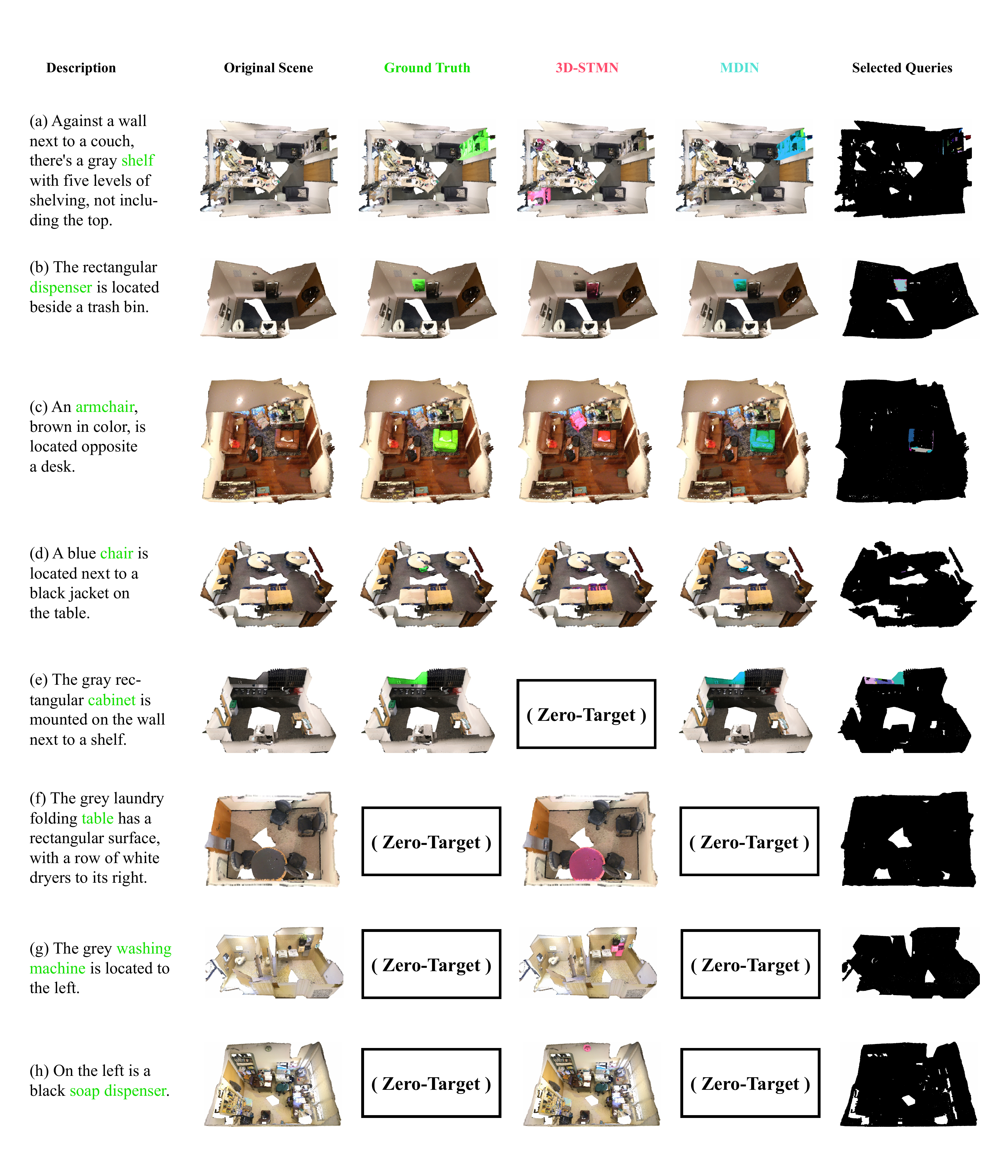}
    \caption{Qualitative comparison between the proposed MDIN and 3D-STMN on single/zero-target cases. Zoom in for the best view.}
    \label{fig:7}
\end{figure*}

More qualitative comparison results of the highly competitive 3D-STMN and MDIN on the Multi3DRes dataset are illustrated in Fig.~\ref{fig:6} and Fig.~\ref{fig:7}.
Fig.~\ref{fig:6} primarily showcases the segmentation results in multi-targets scenarios. 
In multi-targets scenarios, due to the lack of decoupling capability, 3D-STMN either simply segments all instances with the same semantic category as the target (such as \textbf{(a)}, \textbf{(b)}), fails to segment all instances that match the description (such as \textbf{(c)}, \textbf{(d)}), or may even make semantic recognition errors (such as \textbf{(e)}, \textbf{(f)}).
On the contrary, our MDIN accurately segments all instances that match the description. For cases where the target instances have small volumes or complex descriptions (such as \textbf{(g)}, \textbf{(h)}), MDIN also demonstrates its strong discriminative ability. However, 3D-STMN fails to comprehend highly complex language descriptions, leading to erroneous judgments.

Fig.~\ref{fig:7} illustrates the segmentation results for single-target and zero-target scenarios. Benefiting from decoupled modeling, MDIN can capture information about objects in the scene and individually discern their compliance, thus segmenting the correct instances or making accurate predictions when no target instance are present. By contrast, 3D-STMN either suffers from semantic misinterpretation (e.g., \textbf{(a)}, \textbf{(b)}, \textbf{(e)}, \textbf{(g)}, \textbf{(h)}) or makes erroneous segmentation predictions by broadly leveraging semantics (e.g., \textbf{(c)}, \textbf{(d)}, \textbf{(f)}).

We also visualize the superpoints corresponding to the queries selected by the prediction heads of MDIN, as shown in the last column of Fig.~\ref{fig:6} and Fig.~\ref{fig:7}. The results indicate that in the presence of targets, the selected superpoints accurately reflect the positions of the target instances. Notably, for instances with simple geometric features, such as flat surfaces of table, the selected superpoints directly reflect the geometric shapes of the target instances. In the absence of targets, however, no query contains any target instances, hence no superpoints are selected.

\end{document}